\pdfoutput=1

\documentclass[10pt,twocolumn,letterpaper]{article}

\usepackage[pagenumbers]{cvpr}

\usepackage{algorithm}
\usepackage{algpseudocode}
\usepackage{array}
\usepackage{multirow}
\usepackage{makecell}
\usepackage{etoc}
\AtBeginDocument{\etocdepthtag.toc{mainpart}}%
\usepackage{tikz}
\usetikzlibrary{positioning,arrows.meta,calc}
\newcommand{\rs}[1]{{\footnotesize\texttt{#1}}}
\newcommand{\st}[1]{{\footnotesize\bfseries\texttt{#1}}}
\makeatletter
\def\theHALG@line{\thealgorithm.\arabic{ALG@line}}
\makeatother

\definecolor{cvprblue}{rgb}{0.21,0.49,0.74}
\usepackage[pagebackref,breaklinks,colorlinks,allcolors=cvprblue]{hyperref}

\def\confName{CVPR}
\def\confYear{2027}

\title{Code as Tactile Supervisor: Grounding Robot Agents with Demonstration Priors and Tactile Supervision}

\author{%
Zhuoran Li$^{1,2}$\quad
Yizhe Wu$^{2}$\quad
Rui Li$^{2}$\quad
Hao Ma$^{1}$\quad
Yi Ma$^{1}$\\[4pt]
{\normalsize $^{1}$The University of Hong Kong\qquad $^{2}$ViTai Robotics}
}

\makeatletter
\gdef\@title{%
  \begin{minipage}[c]{0.96\textwidth}
    \centering
    Code as Tactile Supervisor%
    \hspace{0.22em}%
    \raisebox{-0.04\height}{%
      \includegraphics[height=0.444cm]{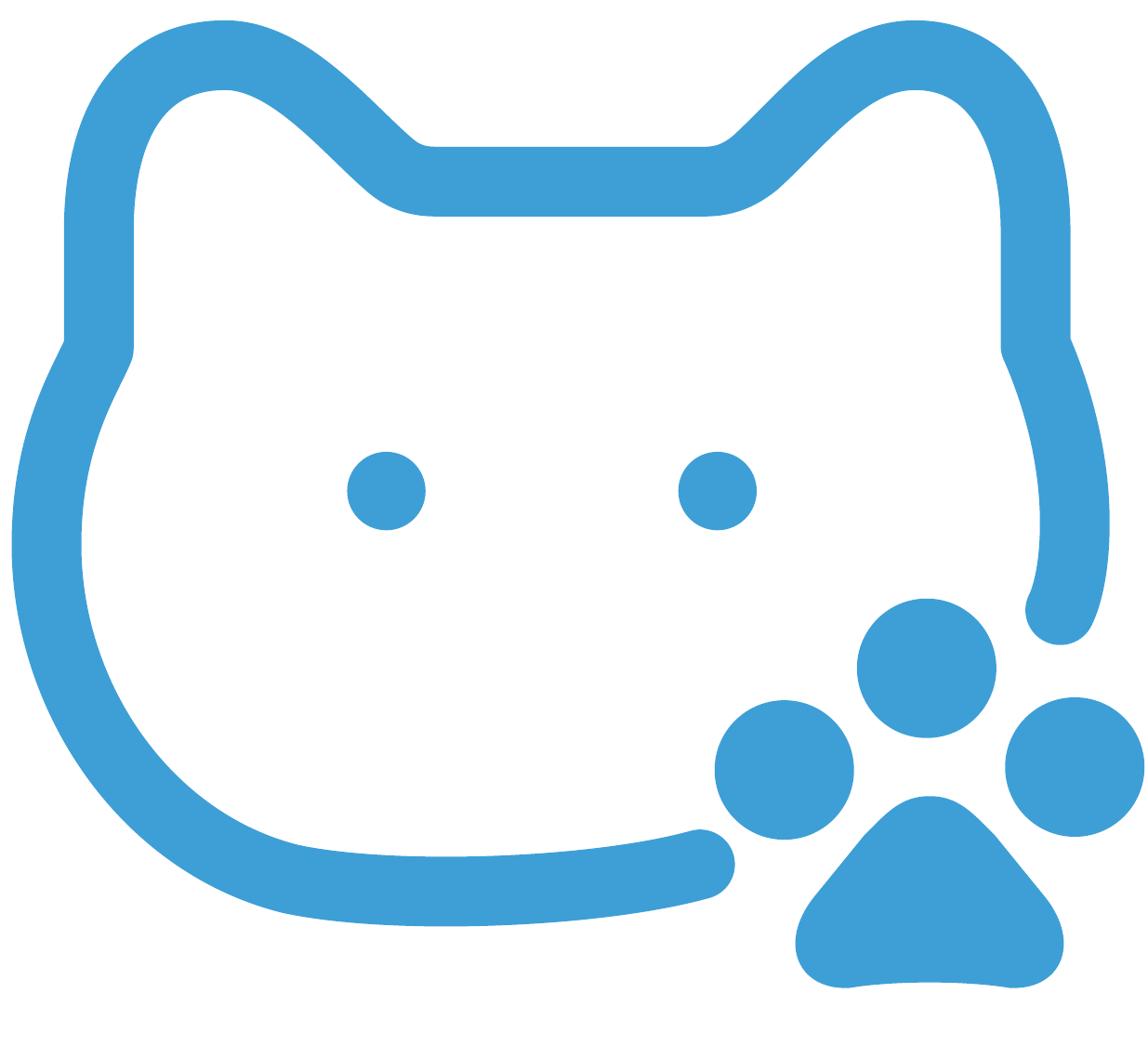}%
    }%
    \hspace{0.12em}: Grounding Robot Agents with\\[-1pt]
    Demonstration Priors and Tactile Supervision
  \end{minipage}%
}
\g@addto@macro\@maketitle{%
  \vspace{-12pt}%
  \begin{minipage}{\textwidth}
    \vspace{-12pt}
    \centering
    \includegraphics[width=\textwidth]{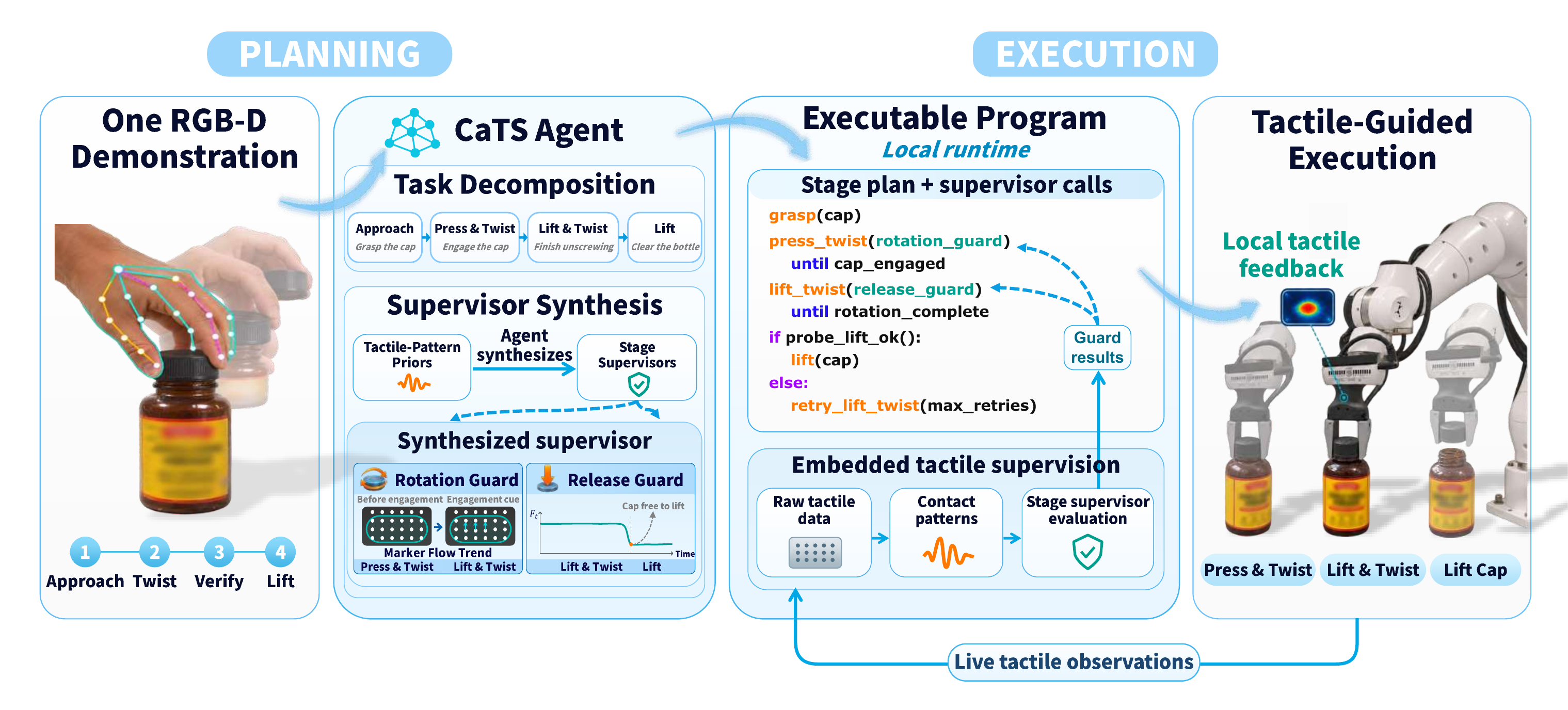}
    \vspace{-16pt}
    \captionof{figure}{\textbf{Overview of CaTS: From one demonstration to a local program with tactile supervision.}
    CaTS adapts a single human RGB-D demonstration into a robot stage plan and uses an agent to synthesize stage-specific supervisors from the demonstration context and tactile-pattern priors, compiling both into a local executable program. In child-resistant cap opening, mechanism priors expand the demonstrated twist into \textit{Press \& Twist} and \textit{Lift \& Twist}; Rotation Guard detects cap engagement and Release Guard verifies release, allowing live tactile feedback to govern these stages and the transition to the final lift.}
    \label{fig:teaser}
  \end{minipage}%
  \vspace{10pt}%
}
\makeatother

\begin{document}
\maketitle
\begin{abstract}
Agentic robot systems can synthesize executable Code-as-Policy programs and revise them from feedback. Their success in simulation depends on accurate state information and inexpensive retries. However, these agents fail in real-world contact-rich manipulation because these advantages are difficult to guarantee. Critical physical states are difficult to observe, and physical trial and error is costly. To address these challenges, we present \underline{C}ode \underline{a}s \underline{T}actile \underline{S}upervisor (\textbf{CaTS}), an agentic framework that grounds robot agents with demonstration priors and tactile supervision. At planning time, CaTS analyzes a single human RGB-D demonstration to recover manipulation stages, motion intent, and key contact relations. Using physics-grounded tactile patterns as priors, the agent synthesizes stage-specific supervisors conditioned on this demonstration-derived task context. CaTS then compiles the robot stage plan and synthesized supervisors into a locally executable program. At execution time, raw tactile observations are encoded into contact patterns, which are evaluated by the synthesized supervisors to monitor contact conditions, drive bounded adjustments, and trigger stage transitions. The program runs locally without further foundation-model calls. Across six real-robot and twenty simulated tasks, CaTS outperforms all evaluated baselines in average success rate. Further evaluations demonstrate cross-sensor transfer and efficient local execution. Together, these results highlight the effectiveness of CaTS in grounding robot agents in real-world contact-rich manipulation.
\end{abstract}
\vspace{-12pt}
\section{Introduction}
\label{sec:intro}

Recent advances in foundation models have enabled agentic robot systems to synthesize executable Code-as-Policy programs~\cite{Liang_2023,Singh_2023,pmlr-v235-mu24a} and revise them from feedback~\cite{Vemprala_2024,shi2026maestro,pmlr-v306-fu26u}. Simulation supports this process with state information and inexpensive retries, making progress easier to assess and alternative strategies easier to test~\cite{pmlr-v306-fu26u}. These advantages are difficult to guarantee in real-world contact-rich manipulation. Vision provides incomplete evidence of contact loads, slip, and hidden mechanism states, so a geometrically plausible motion can fail to produce the intended physical change~\cite{Dong_2021,Lee_2020}. Meanwhile, physical trials require time and resets, limiting the opportunities for open-ended exploration~\cite{eysenbach2018iclr-leave,Ibarz_2021}. These constraints create a gap between simulated success and reliable real-world execution.

Toward reliable real-world execution, prior work has advanced robot programming and perceptual grounding. CaP-X, ASPIRE, and Playful Agentic Robot Learning refine programs through execution feedback, skill synthesis, and experience reuse~\cite{pmlr-v306-fu26u,lu2026aspire,rats2026playful}. However, \textbf{task-specific strategy formation} may still require extensive search, while reused skills may not fit new tasks~\cite{lu2026aspire,rats2026playful}. ReKep~\cite{pmlr-v270-huang25g} and OmniManip~\cite{Pan_2025_CVPR} translate task specifications into geometric constraints, while Code-as-Monitor~\cite{Zhou_2025_CVPR} compiles visual constraints into executable monitors without repeated VLM queries. Yet geometry alone cannot reveal contact loads, slip, or hidden mechanism transitions~\cite{Dong_2021,Lee_2020,Zhou_2025_CVPR}. Recent visuo-tactile agents incorporate tactile observations into action generation~\cite{yang2026native,zhou2026gptgear}, but translating contact evidence into \textbf{motion adjustments and stage transitions} remains challenging; invoking model inference for every new observation also limits \textbf{timely feedback execution}~\cite{yang2026native}.

These challenges motivate three connected questions for real-world contact-rich manipulation. (1) How can an agent form a viable robot plan without privileged simulator state or costly physical exploration? (2) How can raw tactile observations be embedded into a locally executable supervision loop for motion adjustments and stage transitions? (3) How can foundation-model physical priors be leveraged to synthesize this supervision while preserving timely closed-loop responses to tactile feedback?

To address these questions, we present \underline{C}ode \underline{a}s \underline{T}actile \underline{S}upervisor (\textbf{CaTS}), an agentic framework for real-world contact-rich manipulation. CaTS builds on two complementary ideas, \textbf{tactile supervision} and \textbf{demonstration priors}, where demonstration priors guide the synthesis of tactile supervision to ground robot execution in real-world physical interaction. Tactile feedback provides direct evidence of contact states that vision alone cannot reliably observe~\cite{Xue_2025,Wilson_2023,Hogan_2020,Oller_2024}. However, interpreting these signals and selecting suitable responses requires task context~\cite{Su_2016,Li_2026_CVPR}. In turn, a single demonstration supplies this context by illustrating a viable manipulation strategy~\cite{Johns_2021,Wen_2022,Liu_2025,Wu_2024}.

As shown in Fig.~\ref{fig:teaser}, CaTS transforms a single human RGB-D demonstration into a locally executable program with tactile supervision. \textbf{During planning,} CaTS first recovers task structure, motion intent, and key contact relations from the demonstration and adapts them to the current scene and robot to construct a robot stage plan. For each stage, the agent then combines the demonstration-derived task context with physics-grounded tactile-pattern priors to synthesize a stage-specific supervisor. Finally, the agent compiles the stage plan, synthesized supervisors, bounded responses, and transition logic into a local program. \textbf{During execution,} raw tactile observations are encoded into runtime pattern values. At each control step, the fixed active supervisor evaluates these values to assess contact conditions and motion progress, selects a bounded response, and advances the program when its transition condition is satisfied. All feedback processing runs locally without further foundation-model calls.

Through extensive empirical evaluation on six real-robot and twenty simulated tasks, we show that CaTS outperforms all evaluated baselines in average success rate, achieving 66.7\% on the real robot and 61.8\% in simulation. Further evaluations demonstrate cross-sensor transfer and efficient local execution. Our contributions are as follows.
\begin{itemize}[leftmargin=*,nosep]
    \item We introduce \textbf{CaTS, an agentic framework} that grounds robotic agents by combining demonstration priors with tactile supervision.
    \item Hierarchical tactile supervision achieves \textbf{cross-sensor generalizability}, allowing direct deployment of a unified supervision program across heterogeneous tactile sensors without pipeline redesign.
    \item CaTS delivers the \textbf{best overall performance} across twenty simulated and six real-world tasks while substantially reducing cumulative tactile feedback latency.
\end{itemize}

\section{Related Work}
\label{sec:related}

\noindent\textbf{Agentic Programming and Perceptual Grounding.} Language and multimodal models generate robot programs by composing perception, planning, and control APIs~\cite{Liang_2023,pmlr-v235-mu24a,shi2026maestro,Singh_2023,Vemprala_2024}. Building on this programmatic interface, prior work generates reward functions~\cite{ICLR2024_70c26937,ICLR2024_9941833e} and simulation environments~\cite{pmlr-v235-wang24cc}. To improve policies, feedback and self-correction support program refinement, while accumulated experience and demonstrations provide priors for subsequent synthesis~\cite{ICLR2024_2460396f,pmlr-v306-fu26u,liu2026rapid,lu2026aspire,NEURIPS2023_91edff07,NEURIPS2023_1b44b878,wang2024tmlr-voyager,yao2023reactsynergizingreasoningacting,rats2026playful}. Beyond program improvement, spatial constraints and visual tracking connect generated plans to scene geometry~\cite{pmlr-v229-huang23b,pmlr-v270-huang25g,Pan_2025_CVPR}. These geometric representations enable executable monitors to detect constraint violations without repeated model queries during monitoring~\cite{Zhou_2025_CVPR}. However, visual geometry alone does not reveal contact loads, slip, or hidden mechanism transitions~\cite{Dong_2021,Lee_2020,Zhou_2025_CVPR}. Recent agents therefore incorporate tactile observations into action generation~\cite{yang2026native,zhou2026gptgear}, but acting on each new tactile observation still requires another model inference, limiting timely execution.

\noindent\textbf{Tactile Feedback for Manipulation.}
Tactile manipulation follows data- and physics-driven approaches. Data-driven methods learn reusable tactile representations~\cite{pmlr-v270-higuera25a}, map observations to actions through imitation or policy learning~\cite{Xue_2025,liu2025forcemimicforcecentricimitationlearning,Zhu_2025,He_2025}, and inject tactile or force signals into VLA models~\cite{niu2026trextactilereactivedexterousmanipulation,li2026atvlaadaptivetactileinjection,Yu_2025,Li_2026_CVPR}. These task-specific mappings remain difficult to transfer across tasks and objects~\cite{huang2025tactilevlaunlockingvisionlanguageactionmodels,bi2026vlatouch,yuan2026ftp1generalistfoundationtactile}. Physics-driven methods instead map meaningful tactile states and contact modes to explicit closed-loop responses~\cite{Romano_2011,She_2021,Zhao_2025,Su_2016,Wilson_2023,Hogan_2020,Hou_2020,Kim_2022,Oller_2024}, enabling responsive adjustments and stage transitions but requiring task-specific manual definitions of states, conditions, and responses~\cite{Su_2016,Wilson_2023,Hogan_2020,Oller_2024}.

\noindent\textbf{Learning from Few Demonstrations.}
Few demonstrations provide task structure, but effective reuse requires representations that transfer across objects, scenes, and robot embodiments. Prior work adapts demonstrated behavior through visual alignment and object-centered correspondences~\cite{Johns_2021,Wen_2022,Simeonov_2022,pmlr-v229-shen23a,zhu2024densematcherlearning3dsemantic}, while newer methods preserve contact and interaction relations rather than replaying trajectories~\cite{pmlr-v229-biza23a,Liu_2025,pmlr-v100-sieb20a,Zhu_2026}. Decomposition into subgoals and motion primitives~\cite{Zhang_2024,Gao_2024}, together with structured skill representations~\cite{Wu_2024,pmlr-v270-liu25d,pmlr-v270-cheng25a,Nie_2026,zandonati2026rationalinversereasoningfewshot}, supports long-horizon planning and composition. Demonstrations therefore provide strong task and interaction priors, but motion alone cannot reveal when contact-dependent stages are complete, leaving open how these priors can guide stage-specific tactile supervision for local closed-loop execution.

\section{Method}
\label{sec:method}

\noindent\textbf{Problem Setup.} We consider gripper-based contact-rich manipulation. Given a task instruction, a single human RGB-D demonstration, and current scene and robot information, our goal is to synthesize a locally executable program $\Pi$ that enables the robot to complete a complex contact-rich manipulation task in the physical world.

\begin{figure*}[t]
    \centering
    \includegraphics[width=0.95\textwidth]{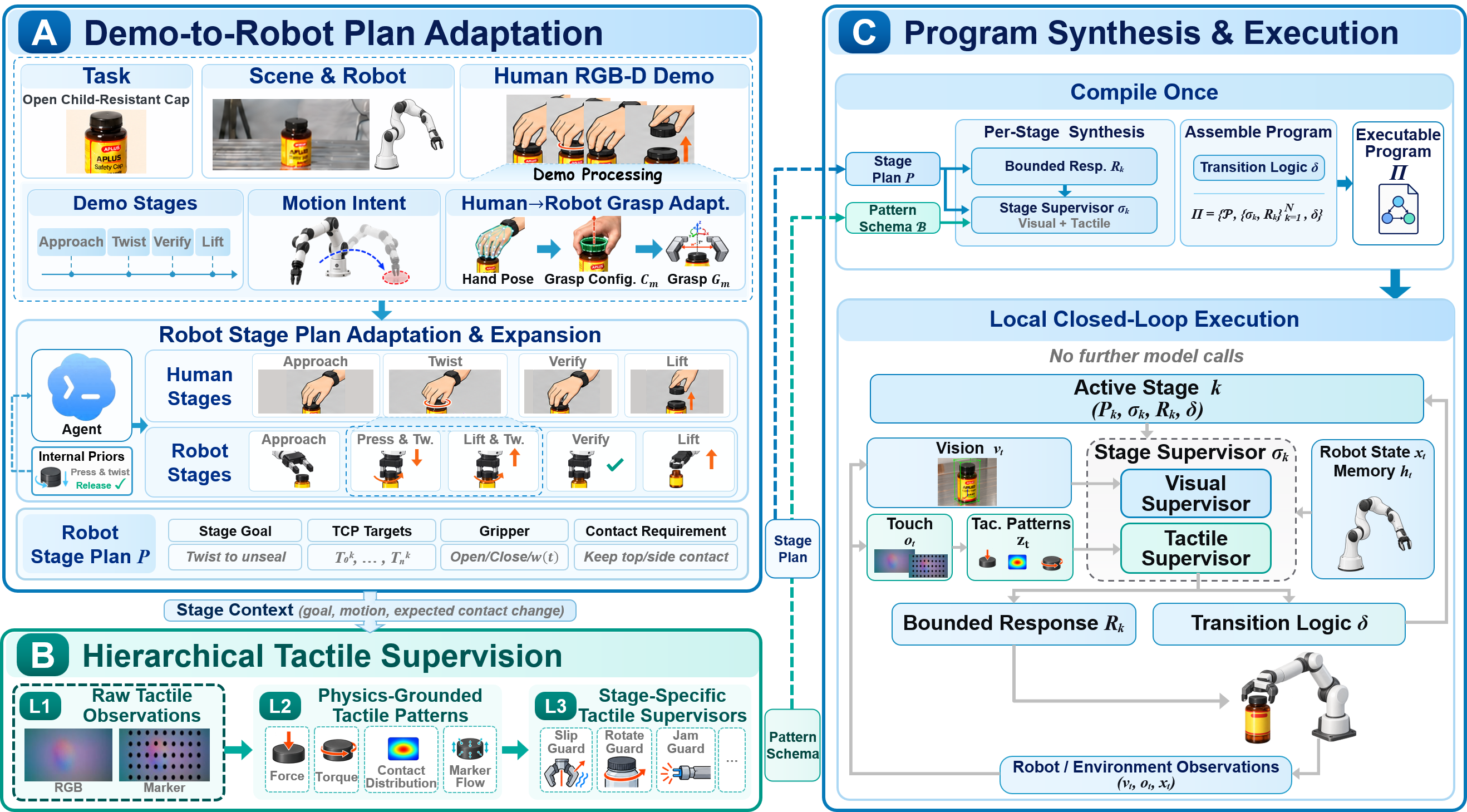}
    \vspace{-4pt}
    \caption{\textbf{CaTS pipeline.} (A) Demo-to-robot plan adaptation converts a single human RGB-D demonstration into a robot stage plan for the current scene and robot. (B) The three-layer tactile hierarchy organizes raw tactile observations, physics-grounded tactile patterns, and stage-specific tactile supervision, linking agent-driven supervisor synthesis during planning with local tactile evaluation during execution. (C) The stage plan and synthesized supervisors are compiled into a local program that governs bounded responses and stage transitions without further foundation-model calls.}
    \label{fig:framework}
    \vspace{-8pt}
\end{figure*}

CaTS combines demonstration priors with tactile supervision to synthesize locally executable robot programs, as illustrated in Fig.~\ref{fig:framework}. CaTS first adapts the human demonstration to the current scene and robot, producing a robot stage plan whose goals, planned motions, and contact requirements carry the demonstration-derived task context (Sec.~\ref{sec:grasp_transfer}). This context connects planning and execution through the three-layer tactile hierarchy (Sec.~\ref{sec:tactile_hierarchy}): during planning, the agent combines it with physics-grounded tactile-pattern priors to synthesize a supervisor for each stage; during execution, the encoder instantiates the corresponding runtime pattern values from raw tactile observations for evaluation by the fixed supervisors. The agent then compiles the stage plan, supervisors, bounded response sets, and transition logic into a local program that closes the feedback loop without further foundation-model calls (Sec.~\ref{sec:policy_synthesis}).

\subsection{Demo-to-Robot Plan Adaptation}
\label{sec:grasp_transfer}

Fig.~\ref{fig:framework}(A) summarizes demo-to-robot plan adaptation.
\hyperref[par:demo_stage_extraction]{Demonstration perception and stage
extraction} first recover an ordered stage sequence, after which
\hyperref[par:grasp_adaptation]{human-to-robot grasp adaptation} derives
robot grasp sets where needed. Finally,
\hyperref[par:stage_expansion]{robot stage plan adaptation and expansion}
combines these outputs with the task instruction, current scene and robot
information, and mechanism priors to construct the robot stage plan.
\vspace{-8pt}

\paragraph{Demonstration perception and stage extraction.}
\label{par:demo_stage_extraction}
CaTS begins by converting the human demonstration into
an ordered stage representation that captures its motion intent and
interaction structure. To obtain this representation, CaTS processes
the object and hand streams in parallel
(Fig.~\ref{fig:framework}(A)). The object branch segments target objects
(SAM3~\cite{carion2026sam3segmentconcepts}) and estimates their 6-DoF poses
(F2G-Pose~\cite{zhang2026f2gposegeometryawarefoundationfeature}) from the
RGB-D stream. In parallel, the hand branch propagates hand masks
(SAM~2.1~\cite{ravi2024sam2segmentimages}), estimates 2D keypoints
(RTMPose~\cite{jiang2023rtmposerealtimemultipersonpose}), and combines them
with local depth to recover 3D hand joints.

CaTS then compares the synchronized hand and object trajectories
to estimate the hand's motion relative to each reference object.
Changes in relative motion and hand--object contact provide geometric
cues for dividing the demonstration into manipulation stages. For
stages in which the grasp is established or reconfigured, the same
trajectories also provide the contact geometry used by the subsequent
grasp-adaptation step.

The agent combines these cues with the demonstration and object roles
to form an ordered stage representation that serves as the
demonstration-derived task context:
\begin{equation*}
    \mathcal D=\{D_m\}_{m=1}^{M},\quad
    D_m=\bigl([t_m^{\mathrm{start}},t_m^{\mathrm{end}}],
    \ell_m,\mu_m,\mathcal I_m\bigr).
\end{equation*}
$M$ is the number of demonstration stages, and
$m\in\{1,\ldots,M\}$ indexes them in time order. Each stage
contains:
\begin{itemize}[leftmargin=*, itemsep=2pt, topsep=2pt]
    \item $[t_m^{\text{start}}, t_m^{\text{end}}]$: \textbf{Temporal interval.} The discrete time steps marking the beginning and end of the stage in the demonstration, where $t_m^{\text{start}}, t_m^{\text{end}} \in \mathbb{Z}_{>0}$ and $t_m^{\text{start}} < t_m^{\text{end}}$.

    \item $\ell_m$:
    \textbf{Stage subgoal.}
    A natural-language description of what the stage aims to achieve.

    \item $\mu_m$:
    \textbf{Motion intent.}
    A natural-language description specifying
    (i) the moving and reference objects,
    (ii) the translation direction or rotation axis, and
    (iii) the intended physical change.

    \item $\mathcal I_m$:
    \textbf{Interaction relations.}
    A natural-language description specifying the participating
    entities and the contacts to be established, maintained, or
    released.
\end{itemize}

For example, a cap-opening demonstration can be divided into four stages:
Approach, Twist, Verify, and Lift. In the Twist stage, $\ell_m$ is
``loosen the cap,'' $\mu_m$ is ``rotate the cap relative to the bottle
about the opening axis,'' and $\mathcal I_m$ specifies ``maintain
hand--cap contact.''

We call demonstration stage $m$ a key-grasp stage if $\mathcal I_m$
indicates that the hand--object grasp is established or reconfigured.
The set of key-grasp stage indices is denoted by
$\mathcal K\subseteq\{1,\ldots,M\}$.
This produces the ordered demonstration stages and
key-grasp indices needed for robot adaptation.

\vspace{-8pt}
\paragraph{Human-to-robot grasp adaptation.}
\label{par:grasp_adaptation}
Using the identified key-grasp stages, CaTS derives
feasible robot grasps from the demonstrated hand--object grasp geometry
in three steps.
\textbf{(1) Extract human grasp constraints.} For each key-grasp stage $m\in\mathcal K$, CaTS selects a short interval in which the demonstrated hand--object grasp is stable. Within this interval, it represents the grasp in the object frame using three geometric cues: the contact location, the closing axis between the opposing fingers, and the approach direction from the wrist toward the contact region. These estimates are aggregated over time to obtain the stage-specific human grasp constraints $\mathcal C_m$. \textbf{(2) Select robot grasps.} Given the current object geometry and robot gripper specifications, including the finger geometry and allowable opening range, GraspGen-X~\cite{Han_2026_CVPR} generates candidate robot grasps. Each candidate is described by the same three geometric quantities. CaTS retains candidates whose contact location, closing axis, and approach direction are consistent with the corresponding constraints in $\mathcal C_m$ and feasible for the current scene and robot embodiment. The retained robot grasps for stage $m$ form the set $\mathcal G_m$. Each grasp specifies an object-relative tool center point (TCP) pose $T_{O\leftarrow G,m}$ and a gripper pre-opening width $w_m$. The selected grasp is denoted by $g_m^{\star}=(T_{O\leftarrow G,m}^{\star},w_m^{\star})$. \textbf{(3) Transform the grasp into the robot base frame.} Camera calibration and the estimated object pose map the selected TCP pose to its execution target, $T_{B\leftarrow G,m}^{\star}=T_{B\leftarrow C}T_{C\leftarrow O}T_{O\leftarrow G,m}^{\star}$.
Here, $B$, $C$, $O$, and $G$ denote the robot base, camera, object, and TCP frames.
The resulting robot grasp sets provide feasible robot
grasps for the corresponding demonstration stages.

\paragraph{Robot stage plan adaptation and expansion.}
\label{par:stage_expansion}
\leavevmode
With the demonstration stages and their corresponding
robot grasp sets in place, the agent combines them with the task instruction,
the current scene and robot information, and mechanism priors to
construct the robot stage plan:
\begin{equation*}
    \overbrace{\left\{D_m\right\}_{m=1}^{M}}^{\mathcal D},
    \quad
    \{\mathcal G_m\}_{m\in\mathcal K}
    \;\longrightarrow\;
    \overbrace{\left\{P_k\right\}_{k=1}^{N}}^{\mathcal P}.
\end{equation*}
Each stage $P_k$ specifies a stage goal, numerical TCP targets, a gripper command, and a contact requirement. TCP targets are represented as either base-frame waypoints or object-relative poses that are updated during execution.

Specifically, the agent adapts the extracted demonstration stages across
three progressive levels:
\begin{itemize}[leftmargin=*, nosep, topsep=3pt]
    \item \textbf{Object Adaptation (Pose):} Maps the demonstrated motion
    to the target object by selecting a feasible equivalent pose allowed
    by symmetry or geometry
    (\eg, choosing a reachable grasp pose from symmetry-equivalent sides
    for an axisymmetric target).
    \item \textbf{Scene Adaptation (Trajectory):} Modifies the motion path
    and inserts waypoints to satisfy obstacle and reachability
    constraints
    (\eg, inserting a detour waypoint to guide the gripper around a newly
    introduced obstacle).
    \item \textbf{Stage Expansion (Task Structure):} Uses mechanism and
    robot priors to split compound actions and add robot-specific
    stages
    (\eg, splitting a demonstrated \textit{Twist} into
    \textit{Press \& Twist} and \textit{Lift \& Twist} to satisfy
    child-resistant cap mechanics and embodiment constraints).
\end{itemize}
Together, these adaptations bridge the gap between human
kinematics and physically grounded robot execution.

The resulting robot stage plan may contain more or fewer stages than the demonstration; therefore, $N$ need not equal $M$.
The resulting plan provides the stage goals, planned motions,
and contact requirements that carry the demonstration-derived task context
into each robot-adapted stage $P_k$. During planning, the agent combines this
stage context with the available tactile-pattern priors to synthesize the
corresponding stage-specific supervisor $\sigma_k$.

\subsection{Hierarchical Tactile Supervision}
\label{sec:tactile_hierarchy}

CaTS structures tactile supervision in three layers (Fig.~\ref{fig:framework}(B)): raw tactile observations (L1), physics-grounded tactile patterns (L2), and the tactile branches of stage-specific supervisors (L3). The middle layer connects planning and execution. \textbf{During planning,} the agent treats the definitions and physical semantics of the L2 patterns as priors and combines them with the demonstration-derived context of each robot stage to synthesize the corresponding L3 tactile branch. \textbf{During execution,} the encoder computes the current L2 pattern values from the L1 observations, and the fixed L3 tactile branch evaluates these values to monitor contact conditions and motion progress, providing evidence for bounded response selection and stage transitions.
\vspace{-8pt}
\paragraph{L1: Runtime raw tactile observations.}
At execution time, the raw observation $o_t$ contains tactile RGB images and marker coordinates, together with sensor identity and timestamp. These observations are inputs to the runtime tactile encoder, not to planning-time supervisor synthesis.

\vspace{-8pt}
\paragraph{L2: Physics-grounded tactile patterns.}
\leavevmode
Each physics-grounded tactile pattern has a fixed physical definition and a runtime value. Let $\mathcal B=\{b_j\}_{j=1}^{J}$ denote this fixed schema of analytical pattern definitions and their physical semantics. During planning, the descriptions in $\mathcal B$ are provided to the agent as tactile-pattern priors. During execution, a unified tactile encoder $E$ instantiates these same pattern types from a window of raw tactile RGB images or marker observations:
\begin{equation*}
    \mathbf z_t = E\!\left(o_{t-W+1:t}\right).
\end{equation*}
Here, $W$ is the observation window length, and $\mathbf z_t$ is a fixed-schema tuple containing the runtime values of the physics-grounded tactile patterns at time $t$. Each value remains an individually accessible scalar, vector, or matrix entry. Table~\ref{tab:tactile_patterns} presents representative pattern types and their runtime representations. Thus, the agent selects pattern types from $\mathcal B$ during planning, whereas $\mathbf z_t$ supplies their current detected values only during execution.

\begin{table}[t]
    \centering
    \footnotesize
    \setlength{\tabcolsep}{1.5pt}
    \renewcommand{\arraystretch}{1.25}
    \caption{Representative physics-grounded tactile pattern types defined in $\mathcal B$ and their runtime representations in $\mathbf z_t$.}
    \label{tab:tactile_patterns}
    \begin{tabular}{@{}
        >{\centering\arraybackslash}m{0.15\linewidth}
        >{\centering\arraybackslash}m{0.17\linewidth}
        >{\centering\arraybackslash}m{0.59\linewidth}@{}}
        \toprule
        Category & Pattern & Mathematical representation \\
        \midrule
        \multirow[c]{3}{=}{\centering\footnotesize\shortstack[c]{Contact\\load}}
        & Normal & $f_t^n\in\mathbb R$ \\
        \cline{2-3}
        & Shear & $\mathbf f_t^s=(f_t^x,f_t^y)\in\mathbb R^2$ \\
        \cline{2-3}
        & Torque & $\boldsymbol{\tau}_t=(\tau_t^x,\tau_t^y,\tau_t^z)\in\mathbb R^3$ \\
        \midrule
        \multirow[c]{4}{=}{\centering\footnotesize\shortstack[c]{Contact\\geometry}}
        & Area & $A_t\in\mathbb R_{\geq 0}$ \\
        \cline{2-3}
        & Center & $\mathbf c_t=(u_t,v_t)\in\mathbb R^2$ \\
        \cline{2-3}
        & Axis & $\phi_t\in[0,\pi)$ \\
        \cline{2-3}
        & Elong. & $e_t\in[1,\infty)$ \\
        \midrule
        \multirow[c]{2}{=}{\centering\footnotesize\shortstack[c]{Marker\\motion}}
        & Global flow & $\bar{\mathbf d}_t=(\bar d_t^u,\bar d_t^v)\in\mathbb R^2$ \\
        \cline{2-3}
        & Resid. shear & $\mathbf R_t=[\mathbf r_{1,t},\ldots,\mathbf r_{N_m,t}]^\top$
        $\in\mathbb R^{N_m\times 2}$ \\
        \bottomrule
    \end{tabular}
\end{table}

\vspace{-12pt}
\paragraph{L3: Stage-specific tactile supervisors.}
During planning, the agent synthesizes a stage-specific supervisor $\sigma_k$ for each stage $P_k$. Its tactile branch is constructed by selecting, composing, and adapting physics-grounded tactile-pattern priors to match the stage goal, planned motion, and expected contact dynamics, while its visual branch is configured from the target geometry and motion-progress requirements. During execution, $\sigma_k$ remains fixed: its tactile branch evaluates the runtime pattern values in $\mathbf z_t$, while its visual branch evaluates the required visual observations.

For \textit{Press \& Twist}, the agent instantiates the tactile branch of $\sigma_k$ as Rotation Guard using the stage context $P_k$, tactile-pattern schema $\mathcal B$, and bounded response set $\mathcal R_k$. It \textit{selects} net tangential force and global marker flow from $\mathcal B$, \textit{adapts} them to the world frame to account for changing fingertip poses, and \textit{composes} their vertical shifts into a unified cap-engagement cue. At runtime, $\sigma_k$ maps the cue computed from $\mathbf z_t$ to either a bounded response in $\mathcal R_k$ or a stage-completion decision according to the demonstration-derived constraints in $P_k$.

\subsection{Program Synthesis and Local Closed-Loop Execution}
\label{sec:policy_synthesis}

\textbf{During program synthesis,} the agent first adapts the demonstration to the current scene and robot to obtain the robot stage plan $\mathcal P$ (Algorithm~\ref{alg:program_synthesis}, line~\ref{alg:adapt_plan}). For each stage $P_k$, it constructs a finite set $\mathcal R_k$ of bounded executable responses, each specifying a TCP or gripper operation and its parameter bounds, and synthesizes $\sigma_k$ from the stage context, tactile-pattern schema $\mathcal B$, and $\mathcal R_k$ (lines~\ref{alg:stage_loop_begin}--\ref{alg:stage_loop_end}). The resulting supervisor maps runtime contact and progress conditions to a response in $\mathcal R_k$ and identifies transition conditions. Finally, the agent connects the stages through $\delta$ and compiles the local program $\Pi$ (lines~\ref{alg:connect_stages}--\ref{alg:compile_program}):
\begin{equation*}
    \Pi =
    \left\{
    \mathcal P,
    \{(\sigma_k,\mathcal R_k)\}_{k=1}^{N},
    \delta
    \right\}.
\end{equation*}

\begin{algorithm}[t]
\caption{Planning-time supervisor and local-program synthesis.}
\label{alg:program_synthesis}
\begin{algorithmic}[1]
\Require $\mathcal D$, $\{\mathcal G_m\}_{m\in\mathcal K}$, task instruction, scene and robot information, tactile-pattern schema $\mathcal B$
\Ensure Executable program $\Pi$
\Statex \textcolor{cvprblue!60}{$\triangleright$\ \textbf{Step 1: Adapt robot stage plan}}
\State $\mathcal P\gets \texttt{adapt\_plan}(\mathcal D,\{\mathcal G_m\}_{m\in\mathcal K},$ \label{alg:adapt_plan}
\Statex \hspace{\algorithmicindent}$\texttt{instruction},\texttt{scene},\texttt{robot})$
\Statex \textcolor{cvprblue!60}{$\triangleright$\ \textbf{Step 2: Synthesize stage supervisors}}
\For{$k=1,\ldots,|\mathcal P|$}\label{alg:stage_loop_begin}
    \State $\mathcal R_k\gets \texttt{build\_bounded\_responses}(P_k)$
    \State $\sigma_k\gets \texttt{synthesize\_supervisor}(P_k,\mathcal B,\mathcal R_k)$
\EndFor\label{alg:stage_loop_end}
\Statex \textcolor{cvprblue!60}{$\triangleright$\ \textbf{Step 3: Assemble program}}
\State $\delta\gets \texttt{connect\_stages}(\mathcal P,\{\sigma_k\})$ \label{alg:connect_stages}
\State \Return $\texttt{compile\_program}(\mathcal P,\{\sigma_k,\mathcal R_k\},\delta)$ \label{alg:compile_program}
\end{algorithmic}
\end{algorithm}

\textbf{During local closed-loop execution in Fig.~\ref{fig:framework}(C),} the compiled program remains fixed. At each control step, the encoder $E$ converts the current raw tactile observations into runtime pattern values $\mathbf z_t$. The active supervisor $\sigma_k$ evaluates these values together with the required visual observations, robot state, and execution memory, and selects a bounded response from $\mathcal R_k$. The program executes the selected response. When $\sigma_k$ confirms a transition condition, $\delta$ updates the active stage. All feedback processing is local and requires no further foundation-model calls.

\section{Experiments}
\label{sec:experiments}

Our experimental evaluation is organized around four key questions.
\textbf{Q1: End-to-end performance (Sec.~\ref{sec:main_results}).}
How well does CaTS perform on contact-rich tasks from a single demonstration,
relative to the evaluated baselines in simulation and on a real robot?
\textbf{Q2: Demonstration priors (Sec.~\ref{sec:ablation}).}
Do demonstration priors improve the effectiveness of tactile supervision?
\textbf{Q3: Cross-sensor transfer (Sec.~\ref{sec:cross_sensor}).}
Can the same supervision program transfer across tactile sensors?
\textbf{Q4: Local feedback latency (Sec.~\ref{sec:latency}).}
Does local execution reduce cumulative feedback latency compared with repeatedly
invoking the agent backbone to process tactile feedback?
\vspace{-4pt}
\subsection{Experimental Setup}
\label{sec:experimental_setup}
\begin{figure}[b]
    \centering
    \includegraphics[width=\columnwidth]{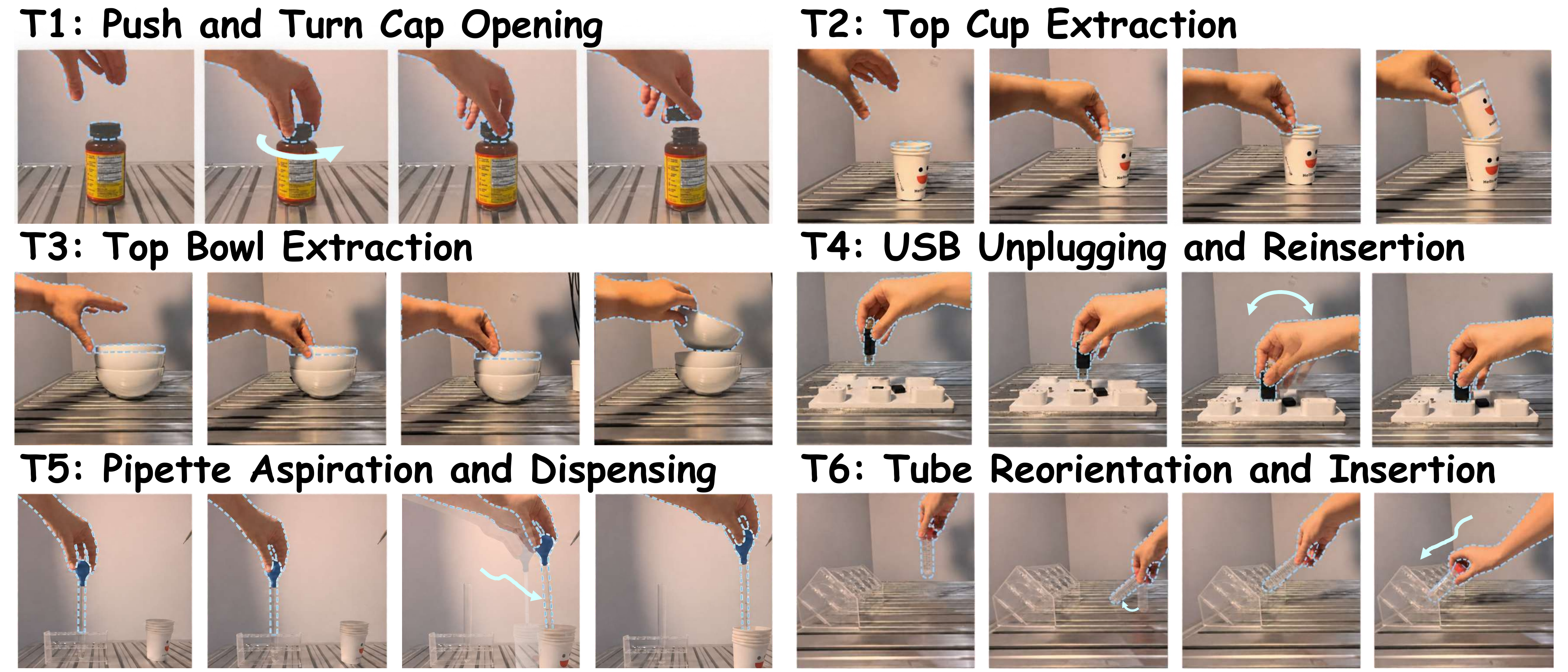}
    \caption{\textbf{Six real-robot tasks and their human demonstrations.}}
    \label{fig:real_robot_tasks}
    \vspace{-8pt}
\end{figure}

\vspace{-4pt}
\paragraph{Real-robot setup.}
We evaluate six contact-rich tasks on a Franka Research 3 arm equipped with
GF220 visuotactile fingertips and an Intel RealSense D435 RGB-D camera.
Fig.~\ref{fig:real_robot_tasks} shows the six tasks and their human
demonstrations. These tasks require complex tactile-guided manipulation whose
contact state and execution progress cannot be reliably determined from vision
alone, with further task details provided in
Appendix~\ref{app:real_robot_tasks}.

\vspace{-4pt}
\paragraph{Simulation setup.}
We further evaluate twenty simulated tasks: eight single-arm tasks from
UniVTAC~\cite{chen2026univtac} and twelve NeoSim~\cite{n0foundation} tasks,
including four single-arm and eight bimanual tasks.
The complete task composition is provided in
Appendix~\ref{app:simulation_benchmarks}.

\vspace{-8pt}
\paragraph{Baselines.}
We compare CaTS with two CaP-Agent0 variants adapted from
CaP-X~\cite{pmlr-v306-fu26u}. For a fair comparison, both variants receive full
raw tactile observations and can autonomously synthesize tactile processing and
control logic. GPT-5.6-sol serves as the shared backbone across all three
methods. The key distinction is therefore the structure provided by the
framework: CaTS explicitly provides physics-grounded tactile patterns and
stage-conditioned supervision, whereas Agent0 operates directly on raw
observations through its native workflow. The two variants are:
\begin{itemize}[leftmargin=*,nosep]
    \item \phantomsection\label{def:agent0}\textbf{Agent0 + Raw Tactile} receives
    raw tactile observations without a demonstration.
    \item \phantomsection\label{def:agent0_demo}\textbf{Agent0 + Raw Tactile +
    Demo} additionally receives the same demonstration as CaTS.
\end{itemize}
We refer to them as \textbf{Agent0} and \textbf{Agent0 + D}, respectively.
The comparison protocol and rationale for baseline selection are provided in
Appendix~\ref{app:comparison_protocol}. Table~\ref{tab:component_comparison}
in the appendix summarizes the component configurations and
comparison targets.

In simulation, we additionally compare with
ACT~\cite{zhao2023learningfinegrainedbimanualmanipulation}, ACT with the
UniVTAC encoder (UniV.)~\cite{chen2026univtac}, and
VITaL~\cite{George_2025} on UniVTAC. On NeoSim~\cite{n0foundation}, we
additionally compare with
$\pi_{0.5}$~\cite{intelligence2025pi05visionlanguageactionmodelopenworld},
LingBot-VA~\cite{li2026causalworldmodelingrobot}, and
Xiaomi-Robotics-0~\cite{cai2026xiaomirobotics0opensourcedvisionlanguageactionmodel}.

\vspace{-8pt}
\paragraph{Evaluation protocol.}
Agent0, Agent0 + D, and CaTS use 10 trials per real-robot task and 90 per
simulated task, with one fixed program per task; we report success rate.
Published baseline results share the official tasks and metric but serve as
reference points, not same-protocol controls.

\subsection{Main Results}
\label{sec:main_results}

\begin{figure}[tb]
    \centering
    \begin{subfigure}{\linewidth}
        \centering
        \includegraphics[width=\linewidth]{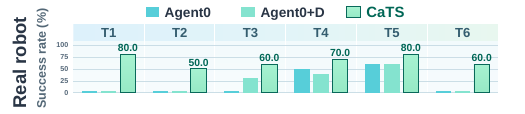}
        \caption{Quantitative comparison with the evaluated baselines.}
        \label{fig:real_robot_success_rates}
    \end{subfigure}
    \vspace{2pt}
    \begin{subfigure}{\linewidth}
        \centering
        \includegraphics[width=\linewidth]{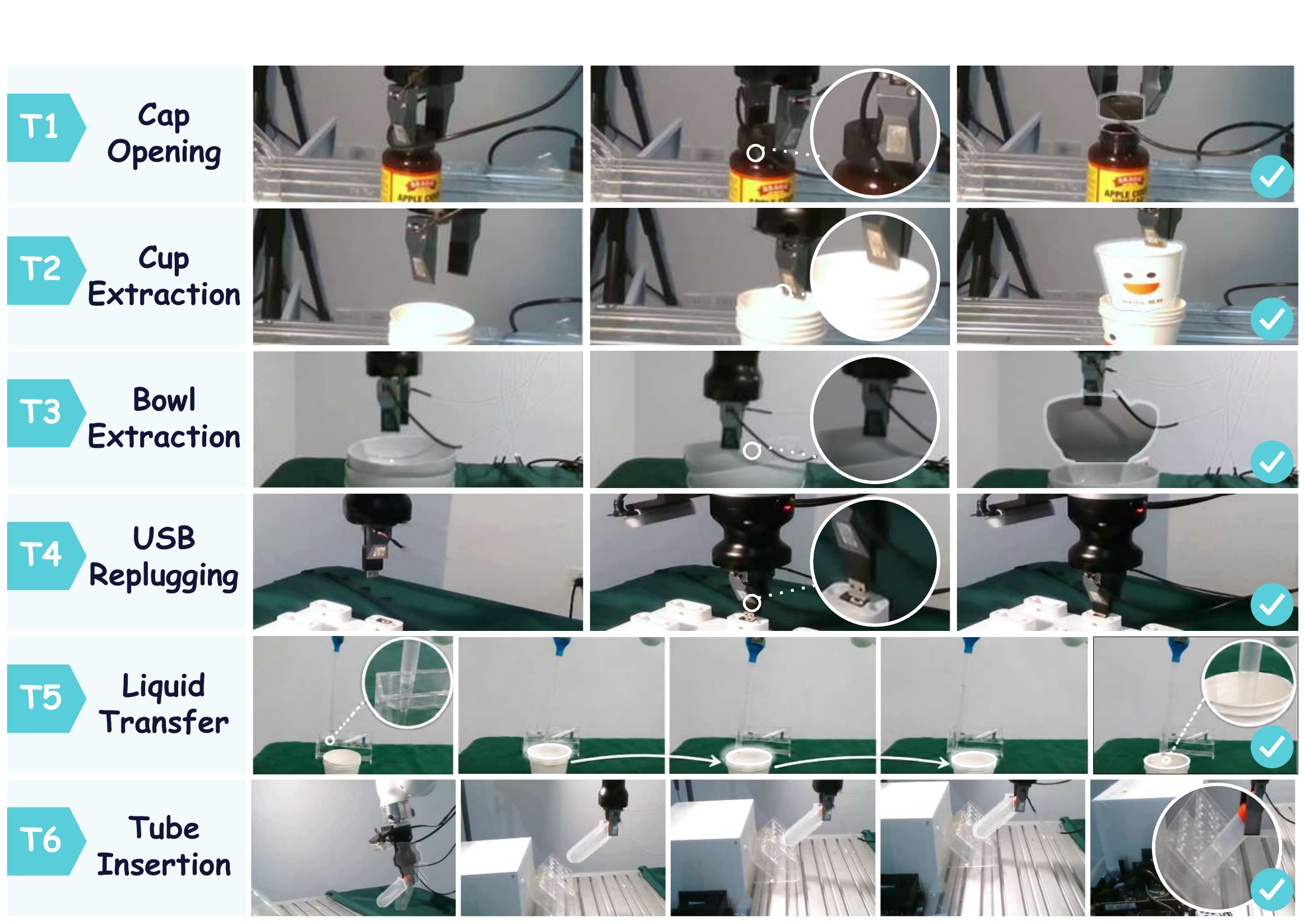}
        \caption{Qualitative execution results of CaTS.}
        \label{fig:real_robot_qualitative}
    \end{subfigure}
    \caption{\textbf{Quantitative and qualitative results on the six real-robot tasks.} (a) Task-level success rates of CaTS and the evaluated baselines. (b) Representative CaTS execution sequences. T1--T6 correspond to the task labels in Fig.~\ref{fig:real_robot_tasks}.}
    \label{fig:real_robot_success}
    \vspace{-8pt}
\end{figure}

\paragraph{Real-robot results.}
CaTS performs best overall on the six real-robot tasks, achieving a mean
success rate of 66.7\%, compared with 21.7\% for Agent0 + D and 18.3\% for
Agent0. Fig.~\ref{fig:real_robot_success}(a) reports the task-level success
rates.

\begin{figure*}[!t]
    \centering
    \includegraphics[width=0.95\textwidth,trim=0 13bp 0 0,clip]{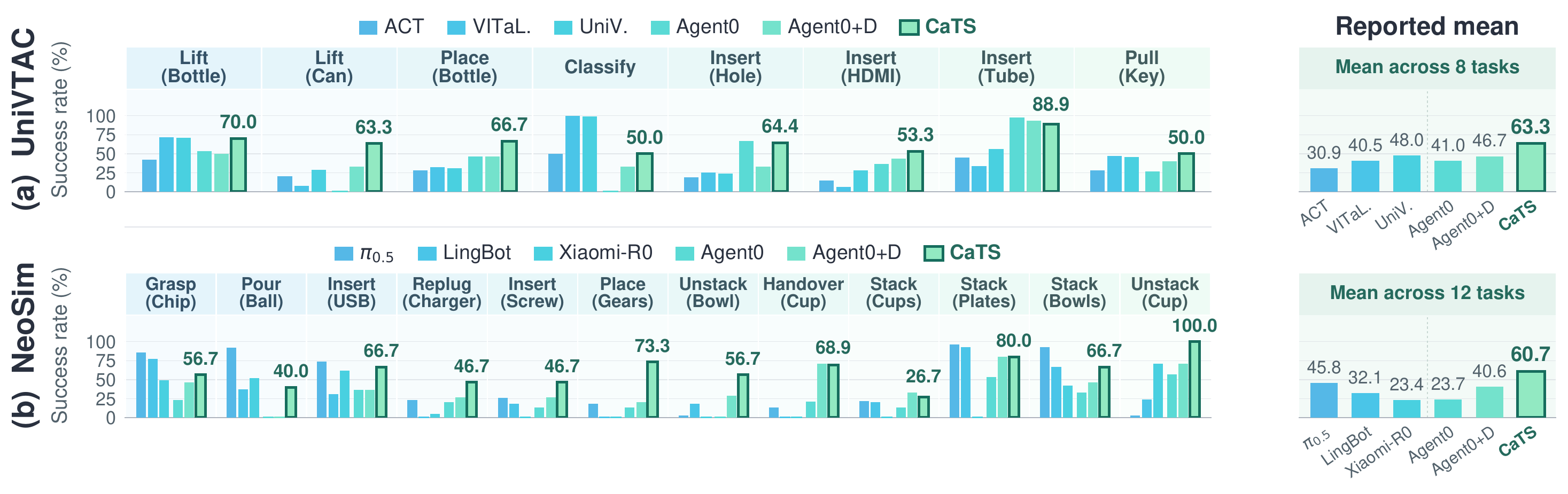}
    \vspace{-8pt}
    \caption{\textbf{Performance comparison on the simulation benchmarks.} Task-level and mean success rates (\%) are reported on (a) UniVTAC~\cite{chen2026univtac} and (b) NeoSim~\cite{n0foundation}.}
    \label{fig:sim_task_results}
    \vspace{-4pt}
\end{figure*}

Both baselines achieve 0\% success on child-resistant cap opening, top-cup extraction, and test-tube reorientation and insertion, while CaTS reaches 80\%, 50\%, and 60\%, respectively. As illustrated by the execution sequences in Fig.~\ref{fig:real_robot_success}(b), these results highlight the roles of three components of CaTS. First, the agentic planning framework leverages the foundation model's physical priors to transform a coarse human demonstration into an executable stage sequence, including the press--twist--release procedure required for cap opening. Second, physics-grounded tactile patterns make subtle raw-sensor changes explicit as task-relevant contact cues, such as the distinction between single-cup and multi-cup contact, allowing the synthesized supervisor to respond reliably during cup extraction. Third, demonstration-to-robot adaptation transfers the effective tube-rotation motion demonstrated by the human, while local tactile supervision monitors rotation progress and applies bounded adjustments until the transparent tube is aligned for insertion, where visual cues alone are unreliable.

\paragraph{Simulation results.}
To assess performance across a broader range of contact-rich manipulation
tasks, we further evaluate CaTS on twenty simulated tasks
(Fig.~\ref{fig:sim_task_results}). CaTS performs best
on both benchmarks, with mean success rates of 63.3\% on UniVTAC and 60.7\% on
NeoSim. Across all twenty tasks, it achieves 61.8\%, compared with 43.1\% for
Agent0 + D and 30.6\% for Agent0.

CaTS is most effective when contact changes must trigger different responses
across stages, as in Insert Hole, Insert Tube, and Pull-out Key. Its strong
performance on Cup Handover, Stack Plates, and Unstack Cup also shows how
demonstrations guide coordinated action modes across the two arms. Together,
these results highlight the benefit of coupling demonstration guidance with
tactile feedback for closed-loop contact-rich execution.

\subsection{Ablation of Demonstration Priors}
\label{sec:ablation}

In the main experiment, comparing CaTS with Agent0 + D (defined in
Sec.~\ref{def:agent0_demo}) demonstrates the advantage of the three-layer
tactile supervision hierarchy over direct access to raw tactile observations.
We further examine whether demonstration priors improve tactile supervision
using an ablation in simulation, as summarized in
Table~\ref{tab:component_comparison} in the appendix.

Across 55 episodes on six representative tasks, we start from full CaTS and
construct \textbf{CaTS w/o Demo Prior} by removing demonstration priors from
tactile supervision while retaining the same robot stage plan and
physics-grounded tactile patterns.
We use next action accuracy to measure whether the appropriate response is
selected at a contact state, and stage success to measure whether the selected
responses complete the current stage through closed loop execution. Together,
the two metrics evaluate tactile supervision at the levels of local decision
making and stage completion. Experimental details are provided in
Appendix~\ref{app:ablation}.

As shown in Table~\ref{tab:stage_ablation}, CaTS achieves 76.9\% next action
accuracy and 80.0\% stage success, compared with 65.1\% and 40.0\% for
CaTS w/o Demo Prior. These results show that demonstration priors make
physics-grounded tactile patterns more effective for closed-loop execution by
organizing them into actionable stage supervision.

\begin{table}[t]
    \centering
    \caption{\textbf{Ablation of demonstration priors in tactile supervision
    across six tasks.} Next action accuracy and stage success are reported for
    CaTS and its ablated variant.}
    \label{tab:stage_ablation}
    \small
    \setlength{\tabcolsep}{5pt}
    \begin{tabular*}{\columnwidth}{@{\extracolsep{\fill}}lcc@{}}
        \toprule
        Method & Next action accuracy & Stage success \\
        \midrule
        CaTS & \textbf{76.9\%} & \textbf{80.0\%} \\
        CaTS w/o Demo Prior & 65.1\% & 40.0\% \\
        \bottomrule
    \end{tabular*}
    \vspace{-8pt}
\end{table}

\subsection{Cross-Sensor Transfer}
\label{sec:cross_sensor}

In simulation, we evaluate cross-sensor transfer on the Lift Can task in
UniVTAC~\cite{chen2026univtac} using GS Mini, GF225, and Xense WS, which
differ in sensing design and raw observations. We configure the supervision
program once on GS Mini and transfer it unchanged to the other sensors by
replacing only the sensor adapter. Three indentation tests and one overload
recovery test, each repeated for 90 trials, assess whether CaTS can map
sensor-specific raw observations to comparable indentation estimates and
preserve the same recovery behavior across sensors. Full protocols are provided
in Appendix~\ref{app:cross_sensor}.

As shown in Table~\ref{tab:cross_sensor_transfer}, the cross sensor difference
in mean slip onset indentation remains below 0.05\,mm in both slip tests, and
the program achieves 100\% overload recovery across sensors.
These results validate the layered abstraction of CaTS: confining sensor
differences to the adapter decouples task supervision from sensor
implementation, enabling deployment of a unified supervision program
across heterogeneous tactile sensors without pipeline redesign.

\begin{table}[t]
    \centering
    \caption{\textbf{Cross-sensor transfer on Lift Can.}
    $D_{\mathrm{slip}}^{\mathrm{eval}}$ and
    $D_{\mathrm{slip}}^{\mathrm{active}}$ denote slip onset under standard and
    induced slip, while $D_{\mathrm{unload}}$ denotes the signed change during
    unloading.}
    \label{tab:cross_sensor_transfer}
    \small
    \setlength{\tabcolsep}{3pt}
    \begin{tabular*}{0.95\columnwidth}{@{\extracolsep{\fill}}lcccc@{}}
        \toprule
        \multirow{2}{*}{Sensor} & \multicolumn{3}{c}{Indentation (mm)} & \multirow{2}{*}{\shortstack{Overload\\recovery}} \\
        \cmidrule(lr){2-4}
        & $D_{\mathrm{slip}}^{\mathrm{eval}}$ & $D_{\mathrm{slip}}^{\mathrm{active}}$ & $D_{\mathrm{unload}}$ & \\
        \midrule
        GS Mini & 5.715 & 5.774 & $-0.384$ & 100\% \\
        GF225 & 5.763 & 5.752 & $-0.970$ & 100\% \\
        Xense & 5.722 & 5.747 & $-0.329$ & 100\% \\
        \bottomrule
    \end{tabular*}
    \vspace{-8pt}
\end{table}

\subsection{Local Feedback Latency}
\label{sec:latency}
\vspace{-4pt}

Tactile stages vary in difficulty and may require different numbers of feedback
decisions. Across five representative tasks in simulation, we compare the
cumulative latency of CaTS and \textbf{CaTS-Online} as these decisions accumulate. CaTS
compiles tactile supervision into a local program before execution, so its
latency includes initial program generation followed by local feedback
decisions. CaTS-Online is a variant that does not perform this compilation;
whenever new tactile feedback arrives, it invokes the shared agent backbone to
interpret the observation and select the next response. Full timing details are
provided in Appendix~\ref{app:latency}.

As shown in Fig.~\ref{fig:latency_amortization}, CaTS recovers its initial
generation cost after approximately 17--25 feedback decisions, after which its
latency advantage grows across five tasks. Thus,
compiling tactile supervision into a local program becomes increasingly
beneficial as feedback demand increases.

\begin{figure}[t]
    \centering
    \includegraphics[width=0.9\columnwidth,trim=0 7bp 0 0,clip]
    {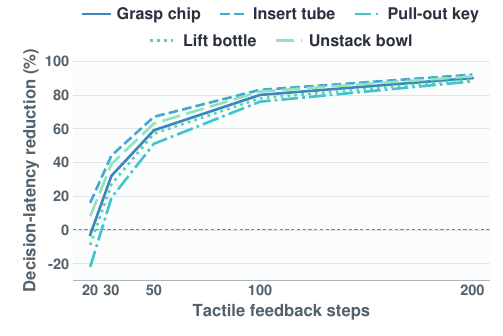}
    \vspace{-8pt}
    \caption{\textbf{Cumulative decision latency of CaTS and CaTS-Online
    across five tasks.} Each curve shows CaTS's relative latency reduction as
    tactile feedback decisions accumulate. The reduction includes the initial
    program generation cost and follows the cumulative-latency formulation in
    Appendix~\ref{app:latency}.}
    \label{fig:latency_amortization}
    \vspace{-16pt}
\end{figure}

\section{Conclusion}
\label{sec:conclusion}
\vspace{-8pt}

We present CaTS, an agentic framework that grounds robot agents with
demonstration priors and tactile supervision. Across six real-robot and twenty
simulated tasks, CaTS achieves the best overall performance among the evaluated
methods. Ablation results demonstrate that demonstration priors improve tactile
supervision. Further evaluations show that the hierarchical tactile supervision
design enables direct transfer of the same supervision program across
heterogeneous tactile sensors, while local execution substantially reduces
cumulative feedback latency. Together, these results demonstrate that CaTS
provides an effective, efficient, and transferable framework for grounding
robot agents in real-world contact-rich manipulation.

\section*{LLM Usage Statement}

LLM tools, including OpenAI (ChatGPT/GPT),
Anthropic (Claude), Google (Gemini), Meta (Llama/Muse),
Alibaba (Qwen), Moonshot (Kimi), and DeepSeek,
were used to obtain reviewer-style feedback on the
manuscript or substantial portions thereof.
These tools provided assistive support for language
editing and refining author-written text, as well as
for code implementation and experimental analysis.
Their involvement in figures, data preparation,
and research ideas, hypotheses, and conclusions
was limited to occasional discussions, suggestions,
and troubleshooting.
All figures, data, research ideas, and scientific
conclusions were independently developed and
finalized by the authors.
No LLM tools were used for preparing related work
or citations. All AI-assisted content was carefully
reviewed and verified by the authors, who take
full responsibility for the submitted work.

{
    \small
    \bibliographystyle{ieeenat_fullname}
    \bibliography{references_strings,references_demo,references_tactile,references_agentic}
}

\twocolumn[{\centering\Large\textbf{Appendix}\par}\vspace{1.0em}]
\appendix
\raggedbottom

\etocdepthtag.toc{appendix}
\etocsettagdepth{appendix}{subsection}
\etocsettagdepth{mainpart}{none}
\makeatletter
\def\apptoc@strip\hskip -1em.~#1\apptoc@end{#1}
\newcommand{\apptocname}{\expandafter\apptoc@strip\etocthename\apptoc@end}
\makeatother
\etocsetstyle{section}{}{}
  {\par\addvspace{3pt}\noindent
   \etoclink{\textbf{\etocthenumber.\enspace\apptocname}}\hfill
   \etoclink{\textbf{\etocthepage}}\par}
  {}
\etocsetstyle{subsection}{}{}
  {\par\noindent\hangindent=3.2em\hspace*{1.2em}%
   \etoclink{\makebox[2em][l]{\etocthenumber}\apptocname}%
   \nobreak\leaders\hbox to 0.6em{\hss.\hss}\hfill
   \nobreak\etoclink{\etocthepage}\par}
  {}
\etocsettocstyle{\section*{Contents}\small\raggedright}{\par\addvspace{6pt}}
\tableofcontents

\setcounter{figure}{0}
\renewcommand{\thefigure}{S\arabic{figure}}
\renewcommand{\theHfigure}{S\arabic{figure}}%

\setcounter{table}{0}
\renewcommand{\thetable}{S\arabic{table}}
\renewcommand{\theHtable}{S\arabic{table}}%

\newcommand{\FtabSetup}{%
    \setlength{\aboverulesep}{0pt}%
    \setlength{\belowrulesep}{0pt}%
    \renewcommand{\arraystretch}{1.3}}
\newcommand{\cmk}{\textcolor{ForestGreen}{$\checkmark$}}
\newcommand{\xmk}{\textcolor{BrickRed}{$\times$}}

\section{Component Configurations}

Agent0 versus CaTS evaluates the end-to-end performance of the complete CaTS
architecture, while Agent0~+~D versus CaTS evaluates the three-layer tactile
supervision hierarchy over direct raw-tactile access.

\paragraph{Two roles of the demonstration prior.}
The demonstration enters CaTS in two places, and the \emph{Demo prior}
column of Table~\ref{tab:component_comparison} refers to a different one in
each group of rows. In the main rows, it denotes the single RGB-D
demonstration provided to the agent, from which the stage plan and motion
strategy are derived; Agent0~+~D and CaTS receive it, and Agent0 does not. In
the ablation rows, it denotes demonstration guidance within tactile
supervision, which constrains tactile response selection inside each stage.
CaTS w/o Demo Prior removes only this guidance and keeps the
demonstration-derived stage plan, so the ablation isolates the
supervision-level role of the demonstration. The stage-level role is
evaluated separately by the Agent0 versus Agent0~+~D comparison in
Appendix~\ref{app:demo_execution_analysis}, and Appendix~\ref{app:ablation}
relates the two results (\emph{Effect of the demonstration on the stage
plan}).

\begin{table}[H]
    \centering
    \caption{\textbf{Component configurations and evaluated contributions.}
    The main comparisons assess end-to-end CaTS and the three-layer tactile
    hierarchy. Both ablation variants retain the same demonstration-derived
    stage plan. \emph{Demo prior} denotes the demonstration provided to the
    agent in the main rows and demonstration guidance within tactile
    supervision in the ablation rows.}
    \label{tab:component_comparison}
    \small\FtabSetup
    \setlength{\tabcolsep}{3.5pt}
    \begin{tabular}{c|l|c|c|c|l}
        \toprule
        & \textbf{Method} &
        \makecell{\textbf{Tactile}\\\textbf{feedback}} &
        \makecell{\textbf{Tactile}\\\textbf{patterns}} &
        \makecell{\textbf{Demo}\\\textbf{prior}} &
        \textbf{Contribution} \\
        \midrule
        \multirow{3}{*}{\rotatebox[origin=c]{90}{\textbf{Main}}}
        & Agent0 & \cmk & \xmk & \xmk & Full CaTS \\
        \cline{2-6}
        & Agent0\,+\,D & \cmk & \xmk & \cmk & \makecell[l]{3-layer\\hierarchy} \\
        \cline{2-6}
        & CaTS & \cmk & \cmk & \cmk & / \\
        \hline
        \multirow{2}{*}{\rotatebox[origin=c]{90}{\textbf{Ablation}}}
        & CaTS\rule[-1.6ex]{0pt}{4.6ex} & \cmk & \cmk & \cmk & / \\
        \cline{2-6}
        & \makecell[l]{CaTS w/o\\Demo Prior} & \cmk & \cmk & \xmk & Demo prior \\
        \bottomrule
    \end{tabular}
\end{table}

\section{Main Evaluation Details}
\label{app:main_setup}

\subsection{Real-Robot Platform and Tasks}
\label{app:real_robot_tasks}

Following the T1--T6 labels in Fig.~\ref{fig:real_robot_tasks}, the six tasks
are described below in terms of their execution procedure and tactile
challenge:
\begin{enumerate}[leftmargin=*, label=\textbf{T\arabic*:}, itemsep=4pt]
    \item \textbf{Child-resistant push-and-turn cap opening.}
    \textbf{Execution.} The robot presses the cap into its engagement slot,
    rotates it until released, and then lifts it from the bottle.
    \textbf{Tactile challenge.} The robot must detect both slot engagement and
    completion of the twist before lifting, while maintaining downward pressure
    and correcting slip during rotation.

    \item \textbf{Top-cup extraction from a nested stack.}
    \textbf{Execution.} The robot probes the stack, adjusts its grasp height,
    and extracts only the uppermost cup.
    \textbf{Tactile challenge.} Subtle contact cues must distinguish a
    single-cup grasp from a grasp involving multiple cups and trigger a grasp
    adjustment when necessary.

    \item \textbf{Top-bowl extraction from a nested stack.}
    \textbf{Execution.} The robot establishes contact, checks the grasp, lifts
    the uppermost bowl, and verifies its separation from the stack.
    \textbf{Tactile challenge.} Because the tactile difference between lifting
    one and multiple nested bowls is weak, the robot must determine whether only
    the top bowl has been isolated before completing the lift.

    \item \textbf{USB unplugging and reinsertion.}
    \textbf{Execution.} The robot removes the USB plug, returns it to the socket,
    and reinserts it.
    \textbf{Tactile challenge.} The gripper occludes the camera near the socket,
    making visual alignment unreliable. Contact feedback therefore guides
    position and orientation corrections during unplugging and reinsertion.

    \item \textbf{Transparent pipette aspiration and dispensing.}
    \textbf{Execution.} The robot aspirates a target amount, follows the cup
    when it is manually displaced, and then dispenses the liquid.
    \textbf{Tactile challenge.} The transparent pipette and liquid provide weak
    visual evidence of transfer progress. Bulb contact and deformation provide
    tactile cues for the aspiration and dispensing stages while the robot
    maintains the relation between the pipette tip and the moving cup.

    \item \textbf{Transparent test-tube reorientation and rack insertion.}
    \textbf{Execution.} The robot rotates the test tube, approaches the rack,
    aligns the tube with the target hole, and inserts it.
    \textbf{Tactile challenge.} Transparency makes the tube pose difficult to
    determine from vision alone. Tactile feedback monitors rotation progress
    and guides position and orientation corrections during contact with the
    rack.
\end{enumerate}
Table~\ref{tab:real_stage_cues} links the stage decisions in T1--T6 to the
visual limitations and tactile evidence available during execution.
Fig.~\ref{fig:real_robot_setup} shows the complete platform and the materials
used in these tasks.

\begin{center}
    \begin{minipage}{\columnwidth}
        \centering
        \includegraphics[width=\columnwidth]{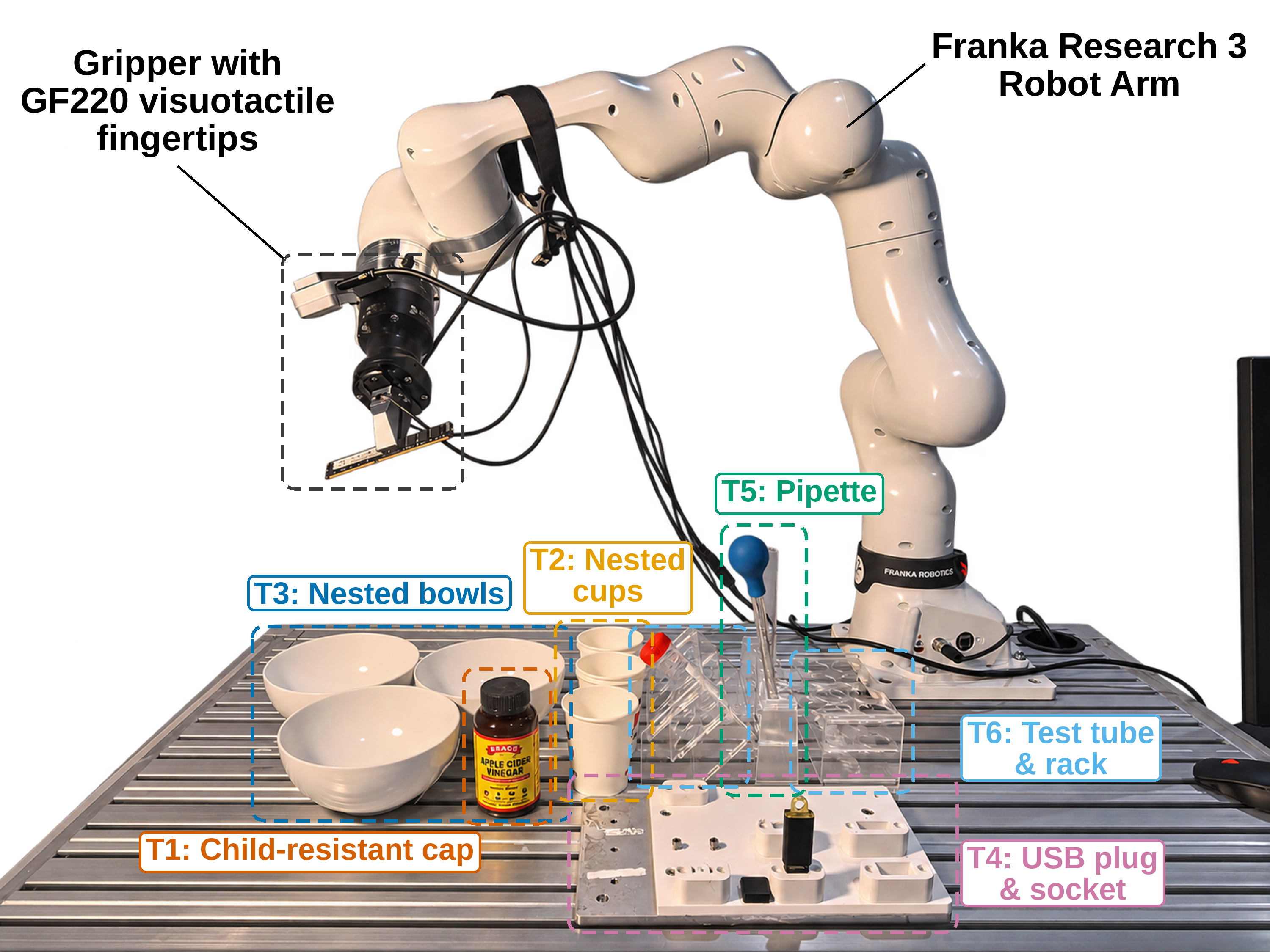}
        \captionof{figure}{\textbf{Real-robot platform and task materials for T1--T6.} Dashed boxes mark the materials used in each task.}
        \label{fig:real_robot_setup}
    \end{minipage}
\end{center}

\begin{table*}[!t]
    \centering
    \caption{\textbf{Stage decisions, visual limitations, and tactile evidence
    in the six real-robot tasks.} Decisions and evidence are listed
    independently; sub-rows do not imply one-to-one correspondence. Tactile
    evidence complements visual geometry, and contact-patch cues support stage
    decisions rather than a direct count of object layers.}
    \label{tab:real_stage_cues}
    \small\FtabSetup
    \setlength{\tabcolsep}{5pt}
    \begin{tabular}{l|l|l|l}
        \toprule
        \textbf{Task} & \textbf{Stage decision} & \textbf{Visual limitation} &
        \textbf{Tactile evidence} \\
        \midrule
        \multirow{2}{*}{\textbf{T1 Cap}}
        & Cap engaged? & \multirow{2}{*}{\makecell[l]{Engagement occurs\\inside the opaque cap}}
        & World-vertical net tangential force and marker-flow shifts \\
        \cline{2-2}\cline{4-4}
        & Cap released? & & Unloading after a probe lift \\
        \hline
        \multirow{2}{*}{\textbf{T2 Cup}}
        & \multirow{2}{*}{Only the top cup grasped?}
        & \multirow{4}{*}{\makecell[l]{Wrist view cannot\\resolve layer count;\\gripper occludes\\frontal view}}
        & Bilateral primary contact-patch number \\
        \cline{4-4}
        & & & Contact-patch area and distribution \\
        \cline{1-2}\cline{4-4}
        \multirow{2}{*}{\textbf{T3 Bowl}}
        & Top bowl alone grasped? & & Contact-patch geometry \\
        \cline{2-2}\cline{4-4}
        & Top bowl separated? & & Load changes during lifting \\
        \hline
        \multirow{2}{*}{\textbf{T4 USB}}
        & Plug aligned? & \multirow{2}{*}{\makecell[l]{Gripper occludes\\the socket region}}
        & Contact cues for bounded pose corrections \\
        \cline{2-2}\cline{4-4}
        & Plug seated? & & Blocking cues \\
        \hline
        \multirow{3}{*}{\textbf{T5 Pipette}}
        & Aspiration stage & \multirow{3}{*}{\makecell[l]{Transparent pipette and\\liquid give weak\\progress cues}}
        & \multirow{2}{*}{Bulb contact changes} \\
        \cline{2-2}
        & Rebound stage & & \\
        \cline{2-2}\cline{4-4}
        & Dispensing stage & & Bulb deformation changes \\
        \hline
        \multirow{2}{*}{\textbf{T6 Tube}}
        & Reorientation progressed? & \multirow{2}{*}{\makecell[l]{Transparency makes\\tube pose tracking unreliable}}
        & Contact changes during rotation \\
        \cline{2-2}\cline{4-4}
        & Insertion progressed? & & Contact changes during rack insertion \\
        \bottomrule
    \end{tabular}
\end{table*}

\paragraph{Per-task analysis.}
The task-level results show how stage-conditioned tactile supervision turns
the same demonstration and raw tactile stream available to Agent0~+~D into
task-specific decisions. On the child-resistant cap, Agent0~+~D succeeds in
0/10 trials and CaTS in 8/10. The coarse demonstration shows cap rotation but
cannot reveal engagement inside the opaque mechanism. The agent uses its
mechanism prior to add press engagement and release verification to the robot
plan. During execution, the joint rise of world-vertical net tangential force
and marker flow gates rotation; insufficient evidence triggers a bounded
additional press while an increment remains. A decrease in
tangential drive and axis moment only proposes release, which a separate probe
lift verifies before the final lift.

Top-cup and top-bowl extraction reach 5/10 and 6/10 with CaTS, compared with
0/10 and 3/10 with Agent0~+~D. During grasp closure, the wrist camera cannot
reliably resolve how many nested layers enter the gripper, while the frontal
view is occluded by the fingers. CaTS converts the tactile images into cues
such as the number, area, and distribution of primary contact patches. These
cues support grasp-height adjustment for the cup and, together with load
changes during lifting, separation verification for the bowl. The cues are
used with stage context rather than as a direct count of grasped layers.

On transparent test-tube reorientation and insertion, CaTS succeeds in 6/10
trials versus 0/10 for Agent0~+~D. The demonstration provides the rotation axis,
direction, and reorientation-before-insertion order. Tactile supervision
monitors rotation progress, determines when insertion should begin, and uses
bounded pose corrections when the tube contacts the rack. This matters when
visual tracking of the transparent tube is unreliable.

\subsection{Simulation Benchmarks}
\label{app:simulation_benchmarks}

\paragraph{UniVTAC.}
Following the official benchmark definitions~\cite{chen2026univtac}, we use
eight single-arm tasks:
\begin{itemize}[leftmargin=*,nosep]
    \item \textbf{Lift Bottle:} grasp a bottle beside a vertical wall and lift
    it while keeping its base close to the wall;
    \item \textbf{Lift Can:} grasp a horizontally placed cylindrical can and
    lift it vertically without slipping;
    \item \textbf{Put Bottle in Shelf:} grasp an upright bottle and place it
    inside a shelf cavity;
    \item \textbf{Grasp Classify:} distinguish visually similar cylinders by
    their surface texture through contact, then place each in its class-specific
    goal region;
    \item \textbf{Insert Hole:} explore the orientation of an inclined hole
    through contact and insert a test tube;
    \item \textbf{Insert HDMI:} align a randomly rotated HDMI connector and
    insert it into a fixed slot;
    \item \textbf{Insert Tube:} insert a test tube through the narrow clearance
    of a hole on an inclined surface;
    \item \textbf{Pull-out Key:} rotate a key until mechanical resistance is
    detected, then pull it straight out of its slot.
\end{itemize}

\paragraph{NeoSim.}
Following the official NeoSim task suite~\cite{n0foundation}, the four
single-arm tasks are:
\begin{itemize}[leftmargin=*,nosep]
    \item \textbf{Grasp Chip:} grasp and lift a thin, fragile chip without
    crushing or dropping it;
    \item \textbf{Pour Ball:} grasp a cup filled with small balls and pour them
    onto a target plate;
    \item \textbf{Plug USB:} align and insert a USB connector into its port;
    \item \textbf{Unplug and Plug Charger:} disconnect a charger connector from
    its socket and insert it again.
\end{itemize}
The eight dual-arm tasks are:
\begin{itemize}[leftmargin=*,nosep]
    \item \textbf{Insert Screw:} stabilize a sleeve with one arm while the other
    inserts and threads a screw into it;
    \item \textbf{Place Gears:} pick up gears and thread them onto their holder
    pegs;
    \item \textbf{Unstack Bowl:} lift the top bowl from a nested stack while the
    other arm stabilizes the remaining bowls;
    \item \textbf{Cup Handover:} transfer a cup from one arm to the other, which
    then places it;
    \item \textbf{Stack Cups:} bring two separated cups together and nest them
    into a centered stack;
    \item \textbf{Stack Plates:} combine two separated plates into a centered
    stack;
    \item \textbf{Stack Bowls:} bring two separated bowls together into a
    centered nested arrangement;
    \item \textbf{Unstack Cup:} remove the top cup from a nested pair while the
    other arm stabilizes the base cup.
\end{itemize}
Together, these tasks cover object reorientation, tactile classification,
insertion, handover, stacking, and separation in both single-arm and bimanual
settings.

\paragraph{Simulation demonstrations.}
For each simulation task, the demonstration is a single RGB-D video of a
successful execution that a human operator produced by teleoperating the
simulated robot in the benchmark scene. The simulators model the robot gripper
rather than a human hand, so the human demonstrates the task through
teleoperation in the embodiment defined by each benchmark, and the
demonstration shares the scene, objects, and camera setup of the evaluation.
The human-hand branch of CaTS, which detects the demonstrating hand and adapts
its grasp to the robot, is evaluated in the real-robot experiments, where all
six tasks use a single human RGB-D demonstration
(Appendix~\ref{app:real_robot_tasks}). In simulation, the teleoperated video is
processed by the same stage-extraction and plan-adaptation pipeline, with the
gripper tracked as the demonstrating effector in place of the hand: the agent
uses the observed object and gripper motion, together with their interaction
relations, to infer stages and adapt the demonstrated strategy to the current
robot scene. Agent0~+~D and CaTS receive the same demonstration video.

Table~\ref{tab:success_summary} summarizes the three agent methods across the
real-robot and simulation evaluations. CaTS succeeds in 40/60 real-robot
trials, versus 13/60 for Agent0~+~D and 11/60 for Agent0. It exceeds Agent0~+~D
on all six real-robot tasks and by 45.0 percentage points in aggregate. Its
mean success rate over twenty simulated tasks is 61.8\%, compared with
43.1\% for Agent0~+~D and 30.6\% for Agent0.

\begin{table}[!t]
    \centering
    \caption{\textbf{Mean task success rates for the three agent methods.}
    Real-robot means cover six tasks; the simulation mean covers the eight
    UniVTAC and twelve NeoSim tasks.}
    \label{tab:success_summary}
    \small\FtabSetup
    \setlength{\tabcolsep}{4pt}
    \begin{tabular}{l|c|c|c|c}
        \toprule
        \textbf{Method} & \textbf{Real} & \textbf{UniVTAC} &
        \textbf{NeoSim} & \textbf{All sim.} \\
        \midrule
        \textbf{Agent0} & 18.3\% & 41.0\% & 23.7\% & 30.6\% \\
        \hline
        \textbf{Agent0~+~D} & 21.7\% & 46.7\% & 40.6\% & 43.1\% \\
        \hline
        \textbf{CaTS} & \textbf{66.7\%} & \textbf{63.3\%} &
        \textbf{60.7\%} & \textbf{61.8\%} \\
        \bottomrule
    \end{tabular}
\end{table}

\subsection{Qualitative Analysis of Demonstration-Guided Execution}
\label{app:demo_execution_analysis}

\begin{figure}[!t]
    \centering
    \includegraphics[width=\columnwidth]{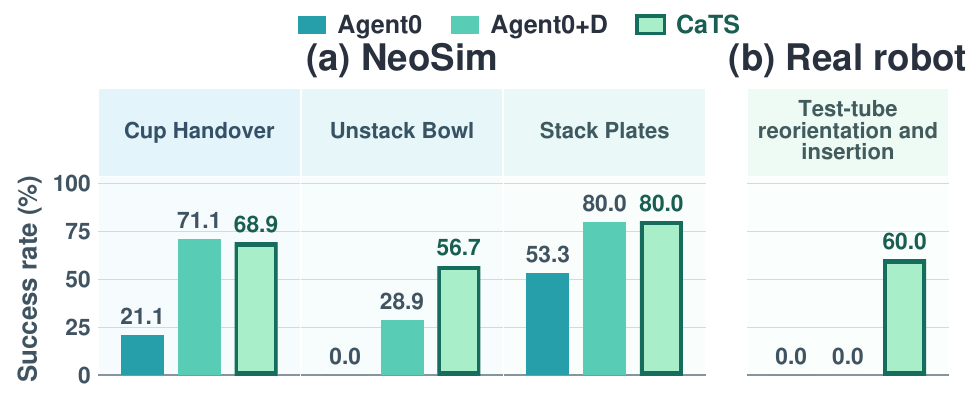}
    \caption{\textbf{Success rates on four demonstration-guided tasks.}
    (a) Three bimanual NeoSim tasks with 90 trials per method and task.
    (b) Real-robot test-tube reorientation and insertion with 10 trials per
    method. Agent0~+~D and CaTS receive the same demonstration.}
    \label{fig:demo_execution_success}
\end{figure}

We analyze four tasks in which successful execution depends on motion and
coordination structure transferred from the demonstration.
Table~\ref{tab:demo_execution_analysis} connects this transferred structure
to the corresponding failure modes observed during execution.
Fig.~\ref{fig:demo_execution_success} compares success rates for the same tasks.

\begin{table*}[!t]
    \centering
    \caption{\textbf{Demonstration-guided execution analysis.} Transferred
    execution structure and corresponding failure modes observed when that
    structure was not recovered. Each row pairs a transferred element with
    the failure observed when it was missing.}
    \label{tab:demo_execution_analysis}
    \small\FtabSetup
    \setlength{\tabcolsep}{5pt}
    \begin{tabular}{l|l|l}
        \toprule
        \textbf{Task} & \textbf{What the demonstration provides} &
        \textbf{Observed failure modes} \\
        \midrule
        \multirow{2}{*}{\makecell[l]{\textbf{Test-tube reorientation}\\\textbf{and insertion}}} & Rotation axis and direction & Wrong rotation axis or direction \\
        \cline{2-3}
         & Reorientation before insertion & Insertion without reorientation \\
        \hline
        \multirow{3}{*}{\makecell[l]{\textbf{Cup Handover}}} & Arm roles & Uncoordinated arm motion \\
        \cline{2-3}
         & Handover sequence & Mistimed handover \\
        \cline{2-3}
         & Release timing & Premature release \\
        \hline
        \multirow{2}{*}{\makecell[l]{\textbf{Unstack Bowl}}} & Stabilizing and manipulating arm roles & Disturbing the remaining stack \\
        \cline{2-3}
         & Stabilization before lifting & Lifting multiple bowls \\
        \hline
        \multirow{2}{*}{\makecell[l]{\textbf{Stack Plates}}} & Relative motion of the two arms & Arm collision \\
        \cline{2-3}
         & Alignment before placement & Off-center stacking \\
        \bottomrule
    \end{tabular}
\end{table*}

Giving Agent0 the demonstration raises success from 21.1\% to 71.1\% on Cup
Handover, from 0\% to 28.9\% on Unstack Bowl, and from 53.3\% to 80.0\% on
Stack Plates. CaTS achieves 68.9\%, 56.7\%, and 80.0\% on these tasks,
respectively. These gains within the same agent framework support the role of
the demonstration in assigning arm roles and sequencing coordinated actions.

On the real-robot test-tube task, CaTS achieves 60.0\% success, compared with
0\% for both Agent0 variants. CaTS adapts the demonstrated rotation axis,
direction, and reorientation-before-insertion sequence to the robot. Its tactile
supervisor monitors rotation progress and determines when to begin insertion.
This example illustrates how the demonstration supplies a motion strategy
that the tactile supervisor helps execute.

\subsection{Comparison Controls and Protocol}
\label{app:comparison_protocol}

All agent-based methods share the same action APIs and access to visual
observations, robot states, and tactile sensing. They also use the same external
visual localization pipeline, initial-state distribution, and success criteria.
Each method generates one program per task, which remains frozen throughout
evaluation.

Each task supplies an instruction and one demonstration. In the real-robot
evaluation, the operator resets the task between trials within the randomized
initial-state distribution. In T5, the operator also moves the cup as specified
by the task protocol. The
shared framework supplies robot APIs and tactile-pattern definitions; the
agent interprets the demonstration, infers any additional mechanism constraints,
and selects stage-specific responses and transitions. Table~\ref{tab:information_sources}
separates these information sources.

\begin{table*}[!t]
    \centering
    \caption{\textbf{Information sources for CaTS program construction.}
    Shared framework inputs are separated from task-specific information and
    decisions made by the agent. A check indicates a task-varying item; a
    cross indicates a fixed input.}
    \label{tab:information_sources}
    \small\FtabSetup
    \setlength{\tabcolsep}{5pt}
    \begin{tabular}{l|l|c}
        \toprule
        \textbf{Source} & \textbf{Information} &
        \makecell{\textbf{Varies by}\\\textbf{task}} \\
        \midrule
        \multirow{3}{*}{\textbf{Shared CaTS framework}}
        & Tactile-pattern definitions & \multirow{3}{*}{\xmk} \\
        \cline{2-2}
        & Tactile-pattern interface & \\
        \cline{2-2}
        & Framework prompt & \\
        \hline
        \textbf{Shared robot interface} & Robot API signatures & \xmk \\
        \hline
        \textbf{Task specification} & Task instruction & \cmk \\
        \hline
        \multirow{3}{*}{\makecell[l]{\textbf{Single RGB-D demonstration}\\
            \textbf{and agent interpretation}}}
        & Stage structure & \multirow{3}{*}{\cmk} \\
        \cline{2-2}
        & Motion intent & \\
        \cline{2-2}
        & Interaction relations & \\
        \hline
        \textbf{Agent inference from the task and mechanism prior}
        & Mechanism constraints & \cmk \\
        \hline
        \multirow{2}{*}{\makecell[l]{\textbf{Agent selection within}\\
            \textbf{permitted robot API ranges}}}
        & Response magnitudes & \multirow{2}{*}{\cmk} \\
        \cline{2-2}
        & Stage-local bounds & \\
        \hline
        \multirow{2}{*}{\makecell[l]{\textbf{Agent synthesis from stage}\\
            \textbf{context and available patterns}}}
        & Tactile event guards & \multirow{2}{*}{\cmk} \\
        \cline{2-2}
        & Stage transitions & \\
        \bottomrule
    \end{tabular}
\end{table*}

\paragraph{Agent0 evaluation scope.}
Both Agent0 variants are evaluated as independent agent frameworks rather than
as ablations of CaTS. We preserve their native iterative code-generation and
refinement workflow and, to provide tactile access, extend only their
observation interface to expose the complete raw tactile stream. Agent0~+~D
additionally receives the same demonstration as CaTS. Incorporating CaTS's
tactile-pattern hierarchy or stage-conditioned supervisors would no longer
evaluate the original Agent0 framework; it would instead constitute an
additional CaTS component ablation. Importantly, Agent0 receives the same robot
action APIs and may freely synthesize intermediate tactile processing, bounded
responses, and transition logic through its native workflow; none of these
capabilities is restricted. Table~\ref{tab:baseline_fairness} summarizes the inputs
and structure available to each method.

\begin{table}[!t]
    \centering
    \caption{\textbf{Fairness controls and method-specific structure.}
    Check marks denote available inputs or permitted capabilities.
    ``Framework-provided'' indicates explicitly supplied structure; crosses do
    not restrict Agent0 from synthesizing equivalent computations or logic.
    All methods execute one frozen program per task locally during evaluation.}
    \label{tab:baseline_fairness}
    \small\FtabSetup
    \setlength{\tabcolsep}{3pt}
    \begin{tabular}{>{\raggedright\arraybackslash}p{0.50\columnwidth}|c|c|c}
        \toprule
        \textbf{Capability} & \textbf{Agent0} & \textbf{Agent0\,+\,D} & \textbf{CaTS} \\
        \midrule
        \textbf{Robot action APIs} & \cmk & \cmk & \cmk \\
        \hline
        \textbf{Vision/localization} & \cmk & \cmk & \cmk \\
        \hline
        \textbf{Raw tactile access} & \cmk & \cmk & \cmk \\
        \hline
        \textbf{Human demonstration} & \xmk & \cmk & \cmk \\
        \hline
        \textbf{May synthesize tactile computations} & \cmk & \cmk & \cmk \\
        \hline
        \textbf{May synthesize transition logic} & \cmk & \cmk & \cmk \\
        \hline
        \textbf{Framework-provided tactile-pattern schema} & \xmk & \xmk & \cmk \\
        \hline
        \textbf{Framework-provided stage-conditioned compilation} & \xmk & \xmk & \cmk \\
        \hline
        \textbf{Local execution of the frozen program} & \cmk & \cmk & \cmk \\
        \bottomrule
    \end{tabular}
\end{table}

\paragraph{Scope of additional agentic baselines.}
We do not directly compare with ASPIRE~\cite{lu2026aspire}, whose
non-prehensile execution assumes an executable motion-primitive interface for
the target environment, or with RAPID~\cite{liu2026rapid}, which relies on
task-specific real-to-sim reconstruction and interactive simulation assets for
program verification and refinement. The corresponding primitives and
validated simulation assets are not available for our evaluated tasks, and
constructing them independently would confound method performance with
task-specific implementation and simulation fidelity.

\paragraph{Scope of tactile-policy baselines.}
Specialized tactile-reactive and visuo-tactile policies are not treated as
interchangeable with Agent0. They typically learn task-specific
observation-to-action mappings from robot trajectories and operate through
method-specific policy and action interfaces. Applying them to the six
real-robot tasks would require task-specific robot data collection, policy
training, and interface adaptation, whereas our setting provides one human
RGB-D demonstration and no task-specific robot training trajectories. This
mismatch is particularly relevant to the heterogeneous, long-horizon,
multi-stage tasks considered here. We therefore use the Agent0 variants as
same-protocol agent baselines for the real-robot evaluation. On UniVTAC, where
published policy results are available under the benchmark protocol, we
additionally include ACT~\cite{zhao2023learningfinegrainedbimanualmanipulation},
ACT with the UniVTAC encoder~\cite{chen2026univtac}, and
VITaL~\cite{George_2025} as published reference points.
On NeoSim, we likewise use results reported by $\mathcal{N}_0$-Foundation~\cite{n0foundation}
for $\pi_{0.5}$~\cite{intelligence2025pi05visionlanguageactionmodelopenworld},
LingBot-VA~\cite{li2026causalworldmodelingrobot}, and
Xiaomi-Robotics-0~\cite{cai2026xiaomirobotics0opensourcedvisionlanguageactionmodel}
as published reference points.

\section{Ablation Details}
\label{app:ablation}

\paragraph{Ablation scope.}
We evaluate the full demo-to-robot pipeline end to end.
Stage extraction, human-to-robot grasp adaptation, and robot stage-plan
adaptation and expansion form the executable robot stage plan.
Appendix~\ref{app:demo_execution_analysis} examines how this plan carries
demonstration-derived task structure into execution.

The tactile-supervision ablation holds the demonstration-derived robot stage
plan and available physics-grounded tactile patterns fixed. It evaluates how
using demonstration guidance to constrain tactile response selection affects
local decisions and stage completion.

\paragraph{Tasks.}
We evaluate six simulation tasks from UniVTAC and NeoSim:
\begin{itemize}[leftmargin=*,nosep]
    \item Grasp Chip;
    \item Grasp Classify;
    \item Lift Bottle;
    \item Insert Tube;
    \item Unstack Bowl;
    \item Pull-out Key.
\end{itemize}

\paragraph{Variants.}
Both settings execute the same robot stage plan and use the same available
physics-grounded tactile patterns. \textbf{CaTS w/o Demo Prior} removes
demonstration guidance from tactile supervision.

\paragraph{Metrics.}
We evaluate tactile supervision at two levels:
\begin{enumerate}[leftmargin=*,nosep,label=(\arabic*)]
    \item \textbf{Next action accuracy:} whether the selected response matches
    the task reference at a stage contact-state decision point;
    \item \textbf{Stage success:} whether the selected responses complete the
    current stage through closed-loop execution.
\end{enumerate}
Both variants are evaluated at the same decision points across 55 episodes.

\paragraph{Choice of metrics.}
The ablation changes only the demonstration guidance within tactile
supervision; the stage plan, tactile patterns, perception, and motion
interfaces are identical across variants. End-to-end task success would
additionally depend on visual localization, grasp acquisition, and stages that
the ablation does not modify, which would dilute the effect of the changed
factor. We therefore evaluate at the contact-state decision points where
tactile supervision acts. Next action accuracy measures the decision itself,
and stage success measures whether those decisions complete the stage in
closed-loop execution, an actual execution outcome rather than an offline
label. Because each task succeeds only when its stages succeed, stage success
captures the part of end-to-end performance that the ablated factor controls.
End-to-end performance of the full system is reported in the main comparison
(Sec.~\ref{sec:main_results}).

\paragraph{Results.}
CaTS reaches 76.9\% next action accuracy, compared with 65.1\% for CaTS w/o
Demo Prior, an 11.8-percentage-point gain in immediate response selection. Its
stage success rises from 40.0\% to 80.0\%, a 40.0-point gain and a doubling of
closed-loop stage completion. Stage success records an actual execution
outcome, so the benefit extends beyond classifying an isolated contact state.
The fixed stage plan and available patterns identify demonstration guidance
within tactile supervision as the changed factor.

\paragraph{Effect of the demonstration on the stage plan.}
This ablation keeps the demonstration-derived stage plan fixed, so it isolates
one of the two roles of the demonstration in CaTS: guiding tactile
supervision. The other role, supplying the stage structure and motion strategy
of the plan, is evaluated by providing the demonstration to the same agent
framework (Appendix~\ref{app:demo_execution_analysis}). Adding the
demonstration to Agent0 raises success from 21.1\% to 71.1\% on Cup Handover,
from 0\% to 28.9\% on Unstack Bowl, and from 53.3\% to 80.0\% on Stack Plates
(Fig.~\ref{fig:demo_execution_success}), and
Table~\ref{tab:demo_execution_analysis} traces the failures without it to
missing arm roles, action sequencing, and alignment before placement. On the
real-robot test-tube task, CaTS adapts the demonstrated rotation axis,
direction, and reorientation-before-insertion order and reaches 60.0\% success,
compared with 0\% for both Agent0 variants. Together, these results and the
ablation cover the stage-level and supervision-level contributions of the
demonstration.

\section{Cross-Sensor Analysis}
\label{app:cross_sensor}

\subsection{Scope and Rationale of Cross-Sensor Evaluation}

Our cross-sensor evaluation examines whether a tactile supervision program
preserves its physical interpretation and corrective behavior when the
underlying sensor changes. We configure the program on GS Mini and replace only
the sensor-specific observation adapter for GF225 and Xense WS. The evaluation
therefore targets portability of the supervision layer across sensor
implementations.

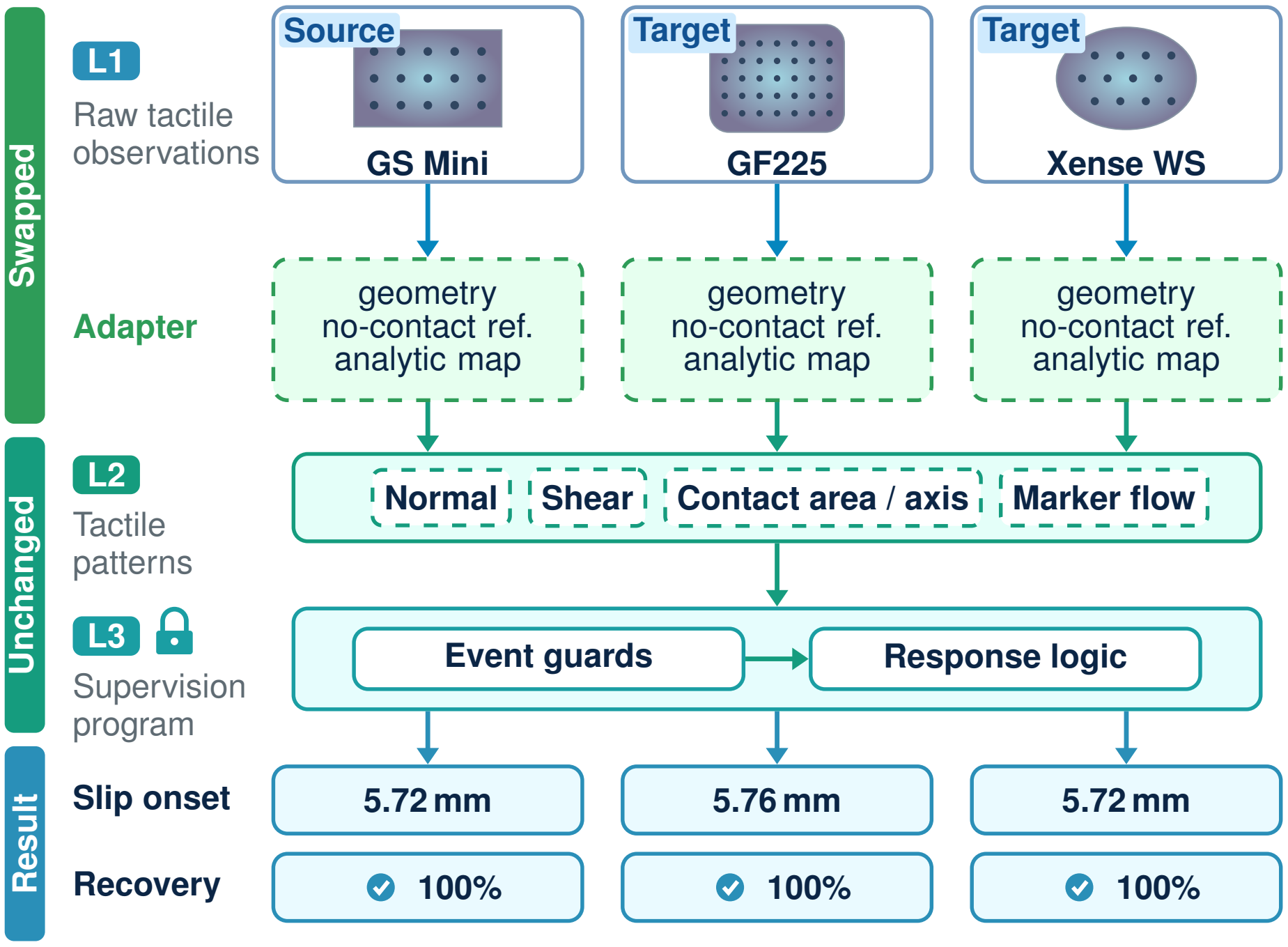
\begin{figure*}[!t]
    \centering
    \resizebox{0.704\textwidth}{!}{%
        \begingroup%
\definecolor{navy}{HTML}{0B2545}%
\definecolor{grayT}{HTML}{5F6B73}%
\definecolor{sky}{HTML}{A9ECFF}%
\definecolor{skyD}{HTML}{2A8FB8}%
\definecolor{cyan}{HTML}{8DF5F8}%
\definecolor{cyanD}{HTML}{1A9EA3}%
\definecolor{aqua}{HTML}{86FDE2}%
\definecolor{aquaD}{HTML}{159C7E}%
\definecolor{mint}{HTML}{9CFFBE}%
\definecolor{mintD}{HTML}{2E9C58}%
\definecolor{tacA}{HTML}{7B7797}%
\definecolor{tacB}{HTML}{9FD3E0}%
\definecolor{blueD}{HTML}{125292}%
\definecolor{blueL}{HTML}{CFEAFD}%
\definecolor{blueA}{HTML}{0586BE}%
\newlength{\lw}\setlength{\lw}{0.8pt}%
\tikzset{%
  font=\sffamily\scriptsize,%
  every node/.style={text=navy, font=\sffamily\scriptsize},%
  box/.style={rounded corners=3pt, line width=\lw},%
  card/.style={box, draw=blueD!60, fill=white},%
  adapter/.style={box, draw=mintD, dashed, fill=mint!25},%
  shared/.style={box, draw=aquaD, fill=aqua!22},%
  frozen/.style={box, draw=cyanD, fill=cyan!22},%
  res/.style={box, draw=skyD, fill=sky!25},%
  sub/.style={box, draw=cyanD, fill=white, inner sep=2pt,%
              font=\sffamily\bfseries\scriptsize},%
  chip/.style={rounded corners=2pt, draw=aquaD, dashed, line width=\lw,%
               fill=white, font=\sffamily\bfseries\scriptsize,%
               minimum height=0.4cm, inner xsep=2.5pt},%
  badge/.style={rounded corners=2pt, fill=#1, text=white,%
                font=\sffamily\bfseries\scriptsize,%
                inner sep=1.6pt, minimum width=0.5cm},%
  flow/.style={-{Triangle[length=3.5pt,width=4.5pt]},%
               line width=1.0pt, color=aquaD},%
  rowlab/.style={anchor=west, align=left,%
                 font=\sffamily\bfseries\scriptsize, inner sep=0pt},%
  rowsub/.style={anchor=north west, align=left,%
                 font=\sffamily\scriptsize, text=grayT, inner sep=0pt},%
}%
\newcommand{\lockicon}[2]{%
  \begin{scope}[shift={#1}, scale=#2]%
    \draw[cyanD, line width=1.0pt] (-0.11,0.02) -- (-0.11,0.14)%
      arc (180:0:0.11) -- (0.11,0.02);%
    \fill[cyanD, rounded corners=0.6pt] (-0.17,-0.17)%
      rectangle (0.17,0.05);%
    \fill[white] (0,-0.05) circle (0.035);%
  \end{scope}}%
\newcommand{\checkicon}[2]{%
  \begin{scope}[shift={#1}, scale=#2]%
    \fill[skyD] (0,0) circle (0.13);%
    \draw[white, line width=0.9pt] (-0.06,0.0) -- (-0.015,-0.05)%
      -- (0.07,0.055);%
  \end{scope}}%
\newcommand{\tacthumb}[3]{%
  \begin{scope}[shift={#1}]%
    \ifcase#2%
      \path[inner color=tacB, outer color=tacA, draw=navy!50,%
            line width=0.3pt] (-0.55,-0.36) rectangle (0.55,0.36);%
    \or%
      \path[inner color=tacB, outer color=tacA, draw=navy!50,%
            line width=0.3pt, rounded corners=4pt]%
            (-0.5,-0.4) rectangle (0.5,0.4);%
    \or%
      \path[inner color=tacB, outer color=tacA, draw=navy!50,%
            line width=0.3pt] (0,0) ellipse (0.52 and 0.38);%
    \fi%
    \ifcase#3%
      \foreach \i in {-2,...,2}{\foreach \j in {-1,...,1}{%
        \fill[navy!85] (\i*0.2,\j*0.2) circle (0.032);}}%
    \or%
      \foreach \i in {-3,...,3}{\foreach \j in {-2,...,2}{%
        \fill[navy!85] (\i*0.13,\j*0.13) circle (0.024);}}%
    \or%
      \foreach \i in {-2,...,1}{\foreach \j in {-1,0,1}{%
        \fill[navy!85]%
          ({\i*0.19+mod(\j+2,2)*0.095+0.05},\j*0.17) circle (0.03);}}%
    \fi%
  \end{scope}}%
\begin{tikzpicture}[x=1cm,y=1cm]
\def\cA{3.15}\def\cB{5.75}\def\cC{8.35}\def\cw{2.3cm}
\def\xl{0.5}\def\xr{9.5}

\fill[mintD, rounded corners=2pt] (0,3.85) rectangle (0.3,6.95);
\node[rotate=90, text=white, font=\sffamily\bfseries\scriptsize]
  at (0.15,5.4) {Swapped};
\fill[aquaD, rounded corners=2pt] (0,1.55) rectangle (0.3,3.75);
\node[rotate=90, text=white, font=\sffamily\bfseries\scriptsize]
  at (0.15,2.65) {Unchanged};
\fill[skyD, rounded corners=2pt] (0,0) rectangle (0.3,1.45);
\node[rotate=90, text=white, font=\sffamily\bfseries\scriptsize]
  at (0.15,0.725) {Result};

\node[badge=skyD, anchor=west] (b1) at (\xl,6.55) {L1};
\node[rowsub] at (\xl,6.25) {Raw tactile\\observations};
\node[rowlab, text=mintD] at (\xl,4.55) {Adapter};
\node[badge=aquaD, anchor=west] at (\xl,3.47) {L2};
\node[rowsub] at (\xl,3.2) {Tactile\\patterns};
\node[badge=cyanD, anchor=west] (b3) at (\xl,2.27) {L3};
\node[rowsub] at (\xl,2.0) {Supervision\\program};
\lockicon{($(b3.east)+(0.25,0)$)}{0.8}
\node[rowlab] at (\xl,1.05) {Slip onset};
\node[rowlab] at (\xl,0.4) {Recovery};

\foreach \x/\i/\name/\shape/\tag in {
  \cA/0/GS Mini/0/Source,
  \cB/1/GF225/1/Target,
  \cC/2/Xense WS/2/Target}{
  \node[card, minimum width=\cw, minimum height=1.3cm] (s\i) at (\x,6.3) {};
  \tacthumb{(\x,6.42)}{\shape}{\shape}
  \node[font=\sffamily\bfseries\scriptsize, anchor=south, inner sep=1.5pt]
    at (s\i.south) {\name};
  \node[rounded corners=1pt, fill=blueL, text=blueD,
        font=\sffamily\bfseries\scriptsize, inner sep=1pt,
        anchor=north west]
    at ($(s\i.north west)+(0.05,-0.05)$) {\tag};
  \node[adapter, minimum width=\cw, minimum height=1.05cm,
        text width=2.1cm, align=center, inner sep=1pt]
    (a\i) at (\x,4.55)
    {geometry\\[-1pt]no-contact ref.\\[-1pt]analytic map};
  \draw[flow, color=blueA] (s\i.south) -- (a\i.north);
}

\node[shared, minimum width=7.2cm, minimum height=0.65cm] (L2) at (\cB,3.3) {};
\node[chip] (c1) at (3.25,3.3) {Normal};
\node[chip, right=0.12 of c1] (c2) {Shear};
\node[chip, right=0.12 of c2] (c3) {Contact area / axis};
\node[chip, right=0.12 of c3] (c4) {Marker flow};
\foreach \i in {0,1,2}{
  \draw[flow] (a\i.south) -- (a\i.south |- L2.north);}

\node[frozen, minimum width=7.2cm, minimum height=0.75cm] (L3) at (\cB,2.1) {};
\node[sub, minimum width=2.9cm, minimum height=0.45cm]
  (g) at (4.05,2.1) {Event guards};
\node[sub, minimum width=2.9cm, minimum height=0.45cm]
  (rr) at (7.45,2.1) {Response logic};
\draw[flow] (g) -- (rr);
\draw[flow] (L2.south) -- (L3.north);

\foreach \x/\i/\v in {\cA/0/5.72, \cB/1/5.76, \cC/2/5.72}{
  \node[res, minimum width=\cw, minimum height=0.5cm,
        font=\sffamily\bfseries\scriptsize] (p\i) at (\x,1.05) {\v\,mm};
  \node[res, minimum width=\cw, minimum height=0.5cm] (q\i) at (\x,0.4) {};
  \checkicon{($(q\i.center)+(-0.35,0)$)}{0.8}
  \node[font=\sffamily\bfseries\scriptsize, anchor=west]
    at ($(q\i.center)+(-0.2,0)$) {100\%};
  \draw[flow, color=skyD] (p\i.north |- L3.south) -- (p\i.north);
}
\pgfresetboundingbox
\path[use as bounding box] (0,-0.05) rectangle (9.55,7.05);
\end{tikzpicture}
\endgroup%
    }
    \caption{\textbf{Cross-sensor transfer of tactile supervision.}
    Sensor-specific L1 observations and adapters change, while the L2 tactile
    pattern interface and L3 supervision program remain fixed. Slip onset
    values are standard-evaluation means rounded to 0.01\,mm; recovery denotes
    overload recovery.}
    \label{fig:cross_sensor_configuration}
\end{figure*}

Calibrated visuotactile simulation can reproduce contact cues measured on
physical sensors: Taxim simulates GelSight marker displacement under normal and
shear loading~\cite{si2022taxim}, and a slip detector
trained only on SimTac images transferred zero-shot to a physical sensor with
92.06\% accuracy~\cite{zhang2026simtac}. These findings motivate evaluating tactile event logic in controlled simulation.

\paragraph{Choice of evaluation.}
End-to-end task success and controlled event tests answer complementary
questions. End-to-end success jointly depends on visual localization, grasp
acquisition, motion planning, and tactile supervision. Holding the task context
fixed instead isolates whether sensor substitution preserves the event
interpretation and corrective behavior of the tactile supervisor.

\subsection{Sensor Adapters and Transfer Configuration}
\label{app:cross_sensor_configuration}

The three UniVTAC sensors differ in raw observation layout,
surface geometry, camera range, and marker layout. Fig.~\ref{fig:cross_sensor_configuration}
traces each sensor's L1 observations through its geometry, no-contact reference,
and analytical adapter into shared L2 physics-grounded tactile patterns. The L3 event
guards, response logic, and response parameters are configured on GS Mini and
reused unchanged on GF225 and Xense WS. The event guards and response logic
form the tactile branch of the stage supervisor $\sigma_k$
(Sec.~\ref{sec:policy_synthesis}), and the response parameters are the bounds
of the responses in $\mathcal R_k$. Only the sensor-specific adapter and its
reference change during transfer.

\paragraph{Adapter implementation.}
Each sensor adapter is lightweight: it is a fixed analytical mapping with no
learned parameters and requires no training data. Setting it up for a new
sensor requires only the known sensor specifications and no-contact reference
frames.
The simulated tactile interface exposes rendered RGB, depth, marker
observations, and the sensor pose according to the capabilities of each sensor
model. For each sensor, no-contact depth and marker frames establish
a sensor-local reference. Depth changes relative to this reference provide
indentation, normalized contact area, and the contact-patch centroid and axis.
Marker displacement provides planar motion and shear information.

The same analytical operations are used for all three sensors. The adapter
accounts for the known image resolution, active surface dimensions, camera
range, marker layout, and sensor pose when converting native observations into
the shared metric representation. The sensor-specific reference calibration
only compensates for observation geometry and the no-contact baseline.
Target-sensor task outcomes are not used to update the event guard, response
logic, or response parameters. Calibration clips from GF225 and Xense WS are
used only to verify that their adapters satisfy the shared physical interface.

\paragraph{Pattern availability and information boundary.}
All three simulated sensors provide the depth and marker observations required
by the selected features, and the transferred program receives only features
computed from rendered tactile observations. Privileged object-relative pose is
used only by the evaluation harness to verify that an induced event occurred
and that the recovery succeeded. Both follow the general rules in
Appendix~\ref{app:tactile_encoder}.

\subsection{Controlled Event Tests}

All tests use the Lift Can task, and each test is repeated for 90 trials. The tests probe complementary properties
of supervision-layer transfer:
\begin{enumerate}[leftmargin=*,nosep,label=(\arabic*)]
    \item \textbf{Evaluation slip onset:} record indentation when slip occurs
    during standard evaluation to test consistency of the physical quantity
    associated with slip;
    \item \textbf{Induced slip onset:} reduce the grip and rotate the can to
    trigger slip under a controlled intervention and test the same event
    interpretation;
    \item \textbf{Unloading:} record the signed indentation change during
    contact unloading to test preservation of the direction and interpretation
    of physical change;
    \item \textbf{Overload recovery:} test whether the frozen program restores
    normal indentation while maintaining both fingertip contacts, providing a
    behavioral test of the transferred response logic.
\end{enumerate}

\subsection{Results and Interpretation}

Slip onset remains consistent across the three sensors: the maximum difference
is 0.048\,mm under standard evaluation and 0.027\,mm under induced slip. The
unloading magnitude varies, but its sign remains negative for every sensor,
preserving the same physical interpretation. The frozen program also achieves
100\% overload recovery on all three sensors. The overload-recovery result
provides behavioral evidence beyond numerical agreement: after sensor
substitution, the frozen program continues to select an effective corrective
response.

Fig.~\ref{fig:cross_sensor_configuration} visualizes
standard-evaluation slip onset and overload recovery after sensor substitution.
Induced-slip and unloading measurements add complementary checks. Together,
these results show that shared L2 patterns and frozen L3 response logic retain
their physical meaning and corrective behavior across the three sensors.

\section{Latency Analysis}
\label{app:latency}

\begin{table*}[!t]
    \centering
    \caption{\textbf{Absolute feedback decision latency on five simulation
    tasks.} Break-even decisions follow the cumulative comparison below, with units
    converted before calculation.}
    \label{tab:absolute_latency}
    \small\FtabSetup
    \setlength{\tabcolsep}{8pt}
    \begin{tabular}{l|c|c|c|c}
        \toprule
        \textbf{Task} &
        \makecell{\textbf{Program generation}\\$T_{\mathrm{prog}}$ (s)} &
        \makecell{\textbf{CaTS local}\\$t_{\mathrm{local}}$ (ms)} &
        \makecell{\textbf{CaTS-Online}\\$t_{\mathrm{online}}$ (s)} &
        \makecell{\textbf{Break-even}\\\textbf{decisions}} \\
        \midrule
        \textbf{Grasp Chip} & 252.34 & 0.0145 & 12.30 & 21 \\
        \hline
        \textbf{Insert Tube} & 269.67 & 0.0160 & 16.05 & 17 \\
        \hline
        \textbf{Pull-out Key} & 298.37 & 0.0151 & 12.20 & 25 \\
        \hline
        \textbf{Lift Bottle} & 280.32 & 0.0192 & 12.88 & 22 \\
        \hline
        \textbf{Unstack Bowl} & 260.67 & 0.0166 & 14.15 & 19 \\
        \bottomrule
    \end{tabular}
\end{table*}

\paragraph{Tasks.}
We evaluate five representative simulation tasks:
\begin{itemize}[leftmargin=*,nosep]
    \item Grasp Chip;
    \item Insert Tube;
    \item Pull-out Key;
    \item Lift Bottle;
    \item Unstack Bowl.
\end{itemize}

\paragraph{Latency components.}
For each task, we measure:
\begin{enumerate}[leftmargin=*,nosep,label=(\arabic*)]
    \item CaTS program generation time;
    \item CaTS local feedback decision latency;
    \item CaTS-Online inference latency when new tactile feedback is sent to
    the shared agent backbone.
\end{enumerate}
The measurements cover feedback sequences of increasing length from critical
contact stages, reflecting stages with different feedback demands.

\paragraph{Cumulative comparison.}
\begin{itemize}[leftmargin=*,nosep]
    \item \textbf{CaTS:} initial program generation followed by $N$ local
    feedback decisions;
    \item \textbf{CaTS-Online:} $N$ online backbone inferences, one for each
    new tactile feedback observation;
    \item \textbf{Feedback horizons:} $N\in\{20,30,50,100,200\}$.
\end{itemize}
For task $j$ and a feedback horizon of $N$ decisions, the decision latency
reduction plotted in Fig.~\ref{fig:latency_amortization} is
\begin{equation*}
    R_j(N)
    =
    100\left(
    1-
    \frac{T_{\mathrm{prog},j}+N t_{\mathrm{local},j}}
         {N t_{\mathrm{online},j}}
    \right),
\end{equation*}
where $T_{\mathrm{prog},j}$ is the CaTS program generation time,
$t_{\mathrm{local},j}$ is its mean local decision latency, and
$t_{\mathrm{online},j}$ is the mean CaTS-Online inference latency per feedback
decision. Positive $R_j(N)$ indicates that CaTS has lower cumulative latency;
$R_j(N)=0$ is the break-even point.

\paragraph{Results.}
CaTS amortizes its initial generation cost after approximately 17--25 feedback
decisions. Beyond this point, its latency advantage grows consistently across
all five tasks, reaching an 87.8\%--91.6\% reduction at 200 decisions. This
trend follows directly from the different scaling behavior: CaTS pays once for
program generation and then executes locally, whereas CaTS-Online repeatedly
invokes the backbone as new tactile feedback arrives.
Table~\ref{tab:absolute_latency} gives the absolute per-task times. For long
contact stages that repeatedly check engagement, correct the grasp, or decide
whether to transition, every additional feedback decision increases the
advantage of local execution over another backbone call.

\section{Implementation and Reproducibility}
\label{app:implementation}

\subsection{Tactile Encoder and Information Boundary}
\label{app:tactile_encoder}

\paragraph{Analytical tactile encoder.}
The tactile encoder $E$ is an analytical processing pipeline rather than a
learned model and requires no task-specific tactile training data. For the
ViTai GF220 sensor used in the real-robot implementation, tactile RGB
deformation and tracked marker motion are processed into contact-geometry and
marker-motion patterns together with FORCE6D-derived tangential-drive
and axis-moment patterns. These force- and moment-related signals are expressed
in sensor-relative units; we do not assume calibrated Newton or Newton-meter
measurements. The supervisors primarily use their direction and window-relative
changes rather than comparing absolute magnitudes.
The real-robot implementation does not use a depth channel; its
contact-geometry patterns are computed directly from the tactile RGB
deformation. For vision-based tactile sensors such as the GF220, a depth map
can also be reconstructed from the tactile RGB image by photometric
stereo~\cite{yuan2017gelsight}.

\paragraph{Sensor-specific adapter contract.}
Each sensor adapter maps the observations available from its sensor into the
shared tactile-pattern schema. The adapter aligns axis directions, sign
conventions, sensor and task frames, bilateral fingertip conventions, and the
temporal direction of change. It does not require heterogeneous sensors to
produce numerically identical raw observations or absolute force magnitudes.
In simulation, the adapters process the rendered RGB, depth, and marker
channels exposed by the corresponding sensor model, as available. Thus, CaTS
unifies sensors at the level of physically interpretable event semantics, such
as increasing contact, unloading, tangential motion, and bilateral imbalance.

\paragraph{Pattern availability.}
Every sensor used in the reported experiments provides the observations needed
by the tactile patterns selected for that experiment. No missing-pattern
fallback is therefore required in the reported evaluations. We do not
reconstruct unavailable patterns from simulator-internal state.

\paragraph{Simulation information boundary.}
In simulation, the tactile encoder, task compiler, and local supervisor receive
only the rendered sensor observations exposed through the simulated tactile
interface. Simulator contact forces, event labels, object states, and other
privileged task-internal variables are not provided as method inputs.
Privileged state may be used by the evaluation harness only to label a
controlled intervention or verify its outcome; it is never available to the
sensor adapter, tactile supervisor, or response-selection logic. All
agent-based methods receive the same raw simulated tactile observations. Table~\ref{tab:tactile_information_boundary}
summarizes this information boundary for real and simulated execution.

\begin{table*}[!t]
    \centering
    \caption{\textbf{Tactile implementation and information boundary.}
    The tactile encoder is analytical, with no task-specific tactile
    training, and is shared by real and simulated settings. Shared cells
    apply to both settings. No missing patterns occurred for the selected
    patterns in the reported experiments.}
    \label{tab:tactile_information_boundary}
    \small\FtabSetup
    \setlength{\tabcolsep}{8pt}
    \begin{tabular}{l|c|c}
        \toprule
        & \textbf{Real sensor} & \textbf{Simulation} \\
        \midrule
        \multirow{3}{*}{\textbf{Observations}} & Tactile RGB & Rendered tactile RGB \\
        \cline{2-3}
        & Tracked marker motion & Rendered marker motion \\
        \cline{2-3}
        & Not used & Rendered depth, where available \\
        \hline
        \multirow{2}{*}{\textbf{Force/moment}} &
        \multicolumn{2}{c}{FORCE6D-derived signals} \\
        \cline{2-3}
        & \multicolumn{2}{c}{Sensor-relative units} \\
        \hline
        \multirow{4}{*}{\textbf{Cross-sensor alignment}} &
        \multicolumn{2}{c}{Shared axes and signs} \\
        \cline{2-3}
        & \multicolumn{2}{c}{Shared sensor and task frames} \\
        \cline{2-3}
        & \multicolumn{2}{c}{Shared bilateral fingertip convention} \\
        \cline{2-3}
        & \multicolumn{2}{c}{Shared direction-of-change semantics} \\
        \hline
        \textbf{Privileged simulator state} & --- &
        Withheld from encoder, compiler, and supervisor \\
        \bottomrule
    \end{tabular}
\end{table*}

\subsection{A Concrete CaTS Program}
\label{app:concrete_program}

We use child-resistant cap opening to show the concrete form of a compiled
CaTS program. The program follows the definition in
Sec.~\ref{sec:policy_synthesis},
\[
    \Pi=\left\{\mathcal P,\{(\sigma_k,\mathcal R_k)\}_{k=1}^{N},\delta\right\},
\]
where $\mathcal P$ is the stage plan; $\mathcal R_k$ is the bounded response
set of stage $k$, in which every response is a concrete robot operation with
explicit parameters and bounds; $\sigma_k$ is the stage supervisor, which
evaluates runtime tactile patterns, visual observations, robot state, and
execution memory to select a response in $\mathcal R_k$ or confirm a transition
condition; and $\delta$ updates the active stage when such a condition holds.
The example instantiates the common program schema and introduces no
task-specific component outside the framework.
Table~\ref{tab:concrete_components} summarizes each component in this task.

\begin{table*}[t]
    \centering
    \caption{\textbf{CaTS program components in the cap-opening task.}
    The program follows
    $\Pi=\{\mathcal P,\{(\sigma_k,\mathcal R_k)\}_{k=1}^{N},\delta\}$
    in Sec.~\ref{sec:policy_synthesis}. $\to$: selected response;
    $\Rightarrow$: flagged stage transition.}
    \label{tab:concrete_components}
    \small\FtabSetup
    \setlength{\tabcolsep}{5pt}
    \begin{tabular}{c|l|l|l}
        \toprule
        \textbf{Comp.} & \textbf{Executable form} &
        \multicolumn{2}{c}{\textbf{Cap-opening instance}} \\
        \midrule
        $\mathcal P$ & Ordered stage plan &
        \multicolumn{2}{l}{7 stages, \rs{APPROACH} $\to\cdots\to$
            \rs{FINAL\_LIFT} (Table~\ref{tab:cap_stages})} \\
        \hline
        \multirow{3}{*}{$\mathcal R_k$} &
        \multirow{3}{*}{\makecell[l]{Bounded robot operations\\
            (operation, parameter, bound)}}
        & \rs{PRESS\_MORE} & $+1$\,mm along bottle axis, at most twice \\
        \cline{3-4}
        & & \rs{TWIST\_CONTINUE} & $10^\circ$ in demo direction, $\le 60^\circ$ total \\
        \cline{3-4}
        & & \multicolumn{2}{l}{All 14 responses with bounds in
            Table~\ref{tab:cap_responses}} \\
        \hline
        \multirow{4}{*}{$\sigma_k$} &
        \multirow{4}{*}{\makecell[l]{Stage supervisor:\\
            events $\to$ response or transition}}
        & \textit{Inputs} & $\mathbf z_t$, visual obs., robot state, execution memory \\
        \cline{3-4}
        & & \textit{Events} & $e_{\mathrm{press}}$, $e_{\mathrm{twist}}$,
            $e_{\mathrm{imb}}$, $e_{\mathrm{release}}$, $e_{\mathrm{unload}}$
            (Table~\ref{tab:cap_events}) \\
        \cline{3-4}
        & & \textit{Selects} & $\neg e_{\mathrm{press}} \to$ \rs{PRESS\_MORE};\;
            $e_{\mathrm{imb}} \to$ \rs{SLIP\_RECOVERY} \\
        \cline{3-4}
        & & \textit{Flags} & $e_{\mathrm{release}} \Rightarrow$ \rs{RELEASE\_VERIFY} \\
        \hline
        \multirow{2}{*}{$\delta$} & \multirow{2}{*}{Stage update on flagged transition}
        & Release candidate & $\Rightarrow$ \rs{RELEASE\_VERIFY} \\
        \cline{3-4}
        & & Unloading confirmed only & $\Rightarrow$ \rs{FINAL\_LIFT} \\
        \bottomrule
    \end{tabular}
\end{table*}

\begin{table*}[t]
    \centering
    \caption{\textbf{Stage-level instantiation of the compiled cap-opening program.}
    Responses are defined in Table~\ref{tab:cap_responses}. Observations and
    responses are listed independently; rows do not imply one-to-one
    correspondence. A satisfied exit guard advances to the next stage in
    $\mathcal P$ unless a target is given by $\Rightarrow$. $\Delta$: window-relative
    change after \rs{PROBE\_LIFT}.}
    \label{tab:cap_stages}
    \small\FtabSetup
    \setlength{\tabcolsep}{4.5pt}
    \begin{tabular}{l|l|l|l|l}
        \toprule
        \textbf{Stage} & \textbf{Goal} & \textbf{Observations / patterns} &
        \textbf{Responses} & \textbf{Exit guard} \\
        \midrule
        \multirow{3}{*}{\st{APPROACH}} &
        \multirow{3}{*}{\makecell[l]{Reach pre-grasp\\pose; no contact}}
        & Cap pose & \multirow{3}{*}{\rs{APPROACH\_CAP}} &
        \multirow{3}{*}{\makecell[l]{Pre-grasp pose\\reached}} \\
        \cline{3-3}
        & & Bottle axis & & \\
        \cline{3-3}
        & & TCP state & & \\
        \hline
        \multirow{2}{*}{\st{GRASP}} &
        \multirow{2}{*}{\makecell[l]{Stable bilateral\\contact}}
        & Bilateral contact load & \rs{CLOSE\_GRIPPER} &
        \multirow{2}{*}{\makecell[l]{Stable bilateral\\contact}} \\
        \cline{3-4}
        & & Gripper state & \rs{HOLD} & \\
        \hline
        \multirow{3}{*}{\st{PRE\_PRESS\_BASELINE}} &
        \multirow{3}{*}{\makecell[l]{Stationary\\reference window}}
        & Freshness, sync. & \multirow{3}{*}{\rs{HOLD\_INVALID}} &
        \multirow{3}{*}{\makecell[l]{Valid reference\\window}} \\
        \cline{3-3}
        & & FORCE6D variation & & \\
        \cline{3-3}
        & & Robot motion state & & \\
        \hline
        \multirow{6}{*}{\st{PRESS\_AND\_TWIST}} &
        \multirow{6}{*}{\makecell[l]{Planned 2\,mm\\entry press;\\confirm engagement;\\
            twist in demo\\direction}}
        & Net tangential force$^{\dagger}$ & \rs{HOLD\_INVALID} &
        \multirow{6}{*}{\makecell[l]{Release candidate\\
            $\Rightarrow$ \rs{RELEASE\_VERIFY}\\[2pt]
            $\alpha_{\mathrm{wrist}}\ge 60^\circ$\\
            $\Rightarrow$ \rs{WRIST\_RESET\_REGRASP}}} \\
        \cline{3-4}
        & & Global marker flow$^{\dagger}$ & \rs{HOLD\_FOR\_ENGAGEMENT} & \\
        \cline{3-4}
        & & Tangential drive & \rs{PRESS\_MORE} & \\
        \cline{3-4}
        & & Axis moment & \rs{START\_TWIST} & \\
        \cline{3-4}
        & & Left--right load imbalance & \rs{TWIST\_CONTINUE} & \\
        \cline{3-4}
        & & Wrist travel $\alpha_{\mathrm{wrist}}$ & \rs{SLIP\_RECOVERY} & \\
        \hline
        \multirow{3}{*}{\st{WRIST\_RESET\_REGRASP}} &
        \multirow{3}{*}{\makecell[l]{Restore wrist range;\\keep a feasible\\cap grasp}}
        & Unloading trend & \rs{UNLOAD} &
        \multirow{3}{*}{\makecell[l]{Stable contact\\restored $\Rightarrow$\\\quad\rs{PRE\_PRESS\_BASELINE}}} \\
        \cline{3-4}
        & & Bilateral contact & \rs{WRIST\_RESET} & \\
        \cline{3-4}
        & & Gripper, wrist state & \rs{REGRASP} & \\
        \hline
        \multirow{3}{*}{\st{RELEASE\_VERIFY}} &
        \multirow{3}{*}{\makecell[l]{Test cap release\\via probe lift}}
        & $\Delta$ axial reaction & \rs{PROBE\_LIFT} &
        \multirow{3}{*}{\makecell[l]{Confirmed $\Rightarrow$ \rs{FINAL\_LIFT}\\Rejected $\Rightarrow$\\\quad\rs{PRE\_PRESS\_BASELINE}}} \\
        \cline{3-4}
        & & $\Delta$ tangential drive & \multirow{2}{*}{\rs{HOLD}} & \\
        \cline{3-3}
        & & TCP state & & \\
        \hline
        \multirow{3}{*}{\st{FINAL\_LIFT}} &
        \multirow{3}{*}{\makecell[l]{Lift cap clear\\of the bottle}}
        & TCP state & \multirow{3}{*}{\rs{LIFT}} &
        \multirow{3}{*}{\makecell[l]{Target height\\reached}} \\
        \cline{3-3}
        & & Gripper state & & \\
        \cline{3-3}
        & & Contact retention & & \\
        \bottomrule
        \multicolumn{5}{l}{\footnotesize $^{\dagger}$World-vertical shift; jointly
            form the Rotation Guard engagement cue (Sec.~\ref{sec:tactile_hierarchy}).}
    \end{tabular}
\end{table*}

\begin{table*}[t]
    \centering
    \caption{\textbf{Bounded response sets $\mathcal R_k$ in the cap-opening program.}
    Each response is a concrete robot operation with explicit parameters and bounds;
    no separate symbolic-to-command binding is used. Unspecified magnitudes lie
    within the robot API ranges supplied to the compiler.}
    \label{tab:cap_responses}
    \small\FtabSetup
    \setlength{\tabcolsep}{5pt}
    \begin{tabular}{l|l|l|l}
        \toprule
        \textbf{Stage} & \textbf{Response} & \textbf{Robot operation} &
        \textbf{Parameter / bound} \\
        \midrule
        \st{APPROACH} & \rs{APPROACH\_CAP} & TCP move to pre-grasp pose &
            Within allowed workspace and speed \\
        \hline
        \multirow{2}{*}{\st{GRASP}} & \rs{CLOSE\_GRIPPER} & Gripper close &
            Within allowed grip range \\
        \cline{2-4}
        & \rs{HOLD} & Hold current pose & One feedback cycle \\
        \hline
        \st{PRE\_PRESS\_BASELINE} & \rs{HOLD\_INVALID} &
            Hold pose; reacquire tactile window & One cycle; window rules (Table~\ref{tab:cap_events}) \\
        \hline
        \multirow{6}{*}{\st{PRESS\_AND\_TWIST}}
        & \rs{HOLD\_INVALID} & Hold pose; reacquire tactile window &
            One cycle; window rules (Table~\ref{tab:cap_events}) \\
        \cline{2-4}
        & \rs{HOLD\_FOR\_ENGAGEMENT} & Hold after the planned entry press &
            One cycle; no translation \\
        \cline{2-4}
        & \rs{PRESS\_MORE} & TCP translation along bottle axis &
            $+1$\,mm; $n_{\mathrm{press}}\le 2$ \\
        \cline{2-4}
        & \rs{START\_TWIST} & Wrist rotation, demo direction & $10^\circ$ \\
        \cline{2-4}
        & \rs{TWIST\_CONTINUE} & Wrist rotation, demo direction &
            $10^\circ$; $\alpha_{\mathrm{wrist}}\le 60^\circ$ \\
        \cline{2-4}
        & \rs{SLIP\_RECOVERY} & Gripper open, then re-close in place &
            Within allowed grip range \\
        \hline
        \multirow{3}{*}{\st{WRIST\_RESET\_REGRASP}}
        & \rs{UNLOAD} & Gripper open (release cap) & Within allowed grip range \\
        \cline{2-4}
        & \rs{WRIST\_RESET} & Wrist rotation, opposite direction & $\le 60^\circ$ \\
        \cline{2-4}
        & \rs{REGRASP} & Re-position to plan pose; gripper close &
            Within allowed grip range \\
        \hline
        \multirow{2}{*}{\st{RELEASE\_VERIFY}}
        & \rs{PROBE\_LIFT} & TCP translation along bottle axis & $+2$\,mm upward \\
        \cline{2-4}
        & \rs{HOLD} & Hold current pose & One feedback cycle \\
        \hline
        \st{FINAL\_LIFT} & \rs{LIFT} & TCP translation along bottle axis &
            $+30$\,mm upward \\
        \bottomrule
    \end{tabular}
\end{table*}

\begin{table*}[t]
    \centering
    \caption{\textbf{Window-relative tactile events in the cap-opening program.}
    ``Persistent'' changes are measured by the self-normalized window change
    $\Delta_x(\mathcal W_t,\mathcal W_0)$; no absolute force threshold is used.}
    \label{tab:cap_events}
    \small\FtabSetup
    \setlength{\tabcolsep}{5pt}
    \begin{tabular}{c|l|l|l}
        \toprule
        \textbf{Event} & \textbf{Patterns} & \textbf{Compiled condition} &
        \textbf{Used in} \\
        \midrule
        $e_{\mathrm{press}}$ & \makecell[l]{Net tangential force,\\global marker flow
            (world vertical)} & Persistent joint rise vs.\ pre-press window &
            \rs{PRESS\_AND\_TWIST} \\
        \hline
        $e_{\mathrm{twist}}$ & Tangential drive &
            Persistent rise in demo rotation direction & \rs{PRESS\_AND\_TWIST} \\
        \hline
        $e_{\mathrm{imb}}$ & Left--right load imbalance $b_{\mathrm{LR}}$ &
            Persistent rise vs.\ last stable-contact window & \rs{PRESS\_AND\_TWIST} \\
        \hline
        $e_{\mathrm{release}}$ & Tangential drive, axis moment &
            Both drop from values confirmed by $e_{\mathrm{twist}}$ & \rs{PRESS\_AND\_TWIST} \\
        \hline
        $e_{\mathrm{unload}}$ & \makecell[l]{Bilateral axial reaction,\\tangential drive} &
            Return toward reference bands after probe lift & \rs{RELEASE\_VERIFY} \\
        \midrule
        $\mathrm{valid}(\mathcal W_t)$ & Tactile window &
            \makecell[l]{$\ge$4 unique samples, span $\ge$60\,ms,\\
            age $\le$0.6\,s, fingertip skew $\le$0.15\,s} & All tactile stages \\
        \bottomrule
    \end{tabular}
\end{table*}

\paragraph{Compiler input and output schema.}
The compiler prompt provides the task instruction, the demonstration-derived
stage structure, descriptions of the available tactile patterns, robot API
signatures, and allowed motion ranges. The compiler outputs a declarative
program in which every stage specifies its goal, required observations, bounded
response set $\mathcal R_k$, and the supervisor $\sigma_k$ that selects
responses and identifies transitions. Generated stages may reference only the
supplied patterns and APIs.

\paragraph{Stage plan $\mathcal P$.}
The human RGB-D demonstration supplies the coarse cap motion and rotation
direction but does not reveal engagement inside the child-resistant mechanism.
Using the task instruction and mechanism prior, the agent expands this motion
into press engagement, rotation, wrist reset and regrasp when needed, release
verification, and final lift. The compiler also receives the current cap pose
and bottle axis, the robot state and APIs, and the active sensor profile.

Within the allowed control ranges, the agent selects a 2-mm initial press, at
most two additional 1-mm presses, $10^\circ$ twist steps, a $60^\circ$
wrist-travel limit, a 2-mm release-verification lift, and a 30-mm final lift.
The initial press is the planned entry motion of \rs{PRESS\_AND\_TWIST}; all
other motions are bounded responses in $\mathcal R_k$
(Table~\ref{tab:cap_responses}). The resulting stage plan is
\[
\begin{split}
\mathcal P:\quad&
\texttt{APPROACH}\rightarrow\texttt{GRASP}\rightarrow\texttt{PRE\_PRESS\_BASELINE}\\
&\rightarrow\texttt{PRESS\_AND\_TWIST}\\
&\rightarrow[\texttt{WRIST\_RESET\_REGRASP}\\
&\hspace{1.2em}\rightarrow\texttt{PRE\_PRESS\_BASELINE}]^{*}\\
&\rightarrow\texttt{RELEASE\_VERIFY}\rightarrow\texttt{FINAL\_LIFT}.
\end{split}
\]
The wrist-reset branch and the fallback after a rejected release verification
are bounded by finite counters in the execution configuration.
Table~\ref{tab:cap_stages} lists the goal, required observations, responses,
and exit guard of every stage.

\paragraph{Bounded response sets $\mathcal R_k$.}
Table~\ref{tab:cap_responses} lists all 14 responses and their parameter
bounds. Each response is directly executable; for example, \rs{PRESS\_MORE}
is a $+1$-mm TCP translation along the bottle axis, executed at most twice,
so no additional symbolic-to-command mapping is needed. Hold responses
(\rs{HOLD}, \rs{HOLD\_INVALID}, and \rs{HOLD\_FOR\_ENGAGEMENT}) keep the
current pose for one feedback cycle while a fresh tactile window is acquired.
Stage transitions, such as entering \rs{RELEASE\_VERIFY}, are not members of
any $\mathcal R_k$; they are identified by $\sigma_k$ and applied by $\delta$.

\paragraph{Window-relative tactile events.}
Rather than comparing sensor-specific absolute force values, $\sigma_k$
compares the current tactile window with a stage-specific reference window.
Let $\mathcal W_0$ denote the reference window for the current twist segment
and $\mathcal W_t$ the current window. For a tactile pattern $x$, its robust
window statistic is
\[
    m_x(\mathcal W)=\operatorname{median}_{\tau\in\mathcal W}x_\tau .
\]
The reference variation $s_x(\mathcal W_0)$ is estimated from the median
absolute deviation, with observed sample variation used when the reference
window is nearly constant. The self-normalized window change is
\[
    \Delta_x(\mathcal W_t,\mathcal W_0)=
    \frac{m_x(\mathcal W_t)-m_x(\mathcal W_0)}{s_x(\mathcal W_0)+\epsilon}.
\]
Events are detected from the direction, persistence, and cross-pattern
consistency of this change. The press-engagement event $e_{\mathrm{press}}$ is
the Rotation Guard of Sec.~\ref{sec:tactile_hierarchy}: it selects net
tangential force and global marker flow, transforms them to the world frame,
and combines their vertical shifts. Left--right load imbalance is defined as
\[
    b_{\mathrm{LR}}(t)=
    \frac{\left|\lVert\mathbf f_t^L\rVert-\lVert\mathbf f_t^R\rVert\right|}
         {\lVert\mathbf f_t^L\rVert+\lVert\mathbf f_t^R\rVert+\epsilon},
\]
and $e_{\mathrm{imb}}$ fires on its persistent increase relative to the most
recent stable-contact window. Table~\ref{tab:cap_events} summarizes all events
and the tactile-window validity conditions. An invalid window triggers
\rs{HOLD\_INVALID} and cannot advance the stage. After every robot command,
the previous window is consumed, and the next decision requires a fresh
post-command window.

\paragraph{Supervisor of \rs{PRESS\_AND\_TWIST}.}
For the core stage, let $m_t$ contain the required visual observations, robot
state, and execution memory, including a one-cycle stage-entry hold condition
$h_t$. After the planned 2-mm entry press, the supervisor
$\sigma_{\mathrm{twist}}(\mathbf z_t,m_t)$ evaluates its rules from top to
bottom:
\[
\begin{array}{@{}r@{\;\;}l@{}}
\neg\,\mathrm{valid}(\mathcal W_t) & \to\text{\rs{HOLD\_INVALID}}\\
h_t & \to\text{\rs{HOLD\_FOR\_ENGAGEMENT}}\\
\neg e_{\mathrm{press}}\land n_{\mathrm{press}}<2 & \to\text{\rs{PRESS\_MORE}}\\
\neg e_{\mathrm{press}}\land n_{\mathrm{press}}=2 & \Rightarrow\text{end attempt}\\
e_{\mathrm{imb}} & \to\text{\rs{SLIP\_RECOVERY}}\\
e_{\mathrm{release}} & \Rightarrow\text{\rs{RELEASE\_VERIFY}}\\
\alpha_{\mathrm{wrist}}\ge 60^\circ & \Rightarrow\text{\rs{WRIST\_RESET\_REGRASP}}\\
e_{\mathrm{press}}\land n_{\mathrm{twist}}=0 & \to\text{\rs{START\_TWIST}}\\
\text{otherwise} & \to\text{\rs{TWIST\_CONTINUE}}
\end{array}
\]
Here, $\to$ selects a response in $\mathcal R_{\mathrm{twist}}$ and
$\Rightarrow$ flags a transition applied by $\delta$; $n_{\mathrm{press}}$
counts additional presses, $n_{\mathrm{twist}}$ counts executed twist steps,
and $\alpha_{\mathrm{wrist}}$ is the accumulated wrist travel. The condition
$h_t$ is cleared at the first valid window after the entry press. This rule
order ensures that rotation occurs only after $e_{\mathrm{press}}$ holds: if
two additional presses do not confirm engagement, the current rotation attempt
ends.

\paragraph{Independent release verification.}
A release candidate cannot directly activate \rs{FINAL\_LIFT}. It is declared
only when both tangential drive and axis moment persistently decrease from the
elevated values confirmed by $e_{\mathrm{twist}}$:
\[
    e_{\mathrm{release}}=\operatorname{drop}(q_{\mathrm{drive}})\land
    \operatorname{drop}(m_{\mathrm{axis}}).
\]
The program then stops twisting and enters \rs{RELEASE\_VERIFY}: it executes
\rs{PROBE\_LIFT} (2\,mm upward along the bottle axis) and acquires a fresh
tactile window. If the bilateral axial reaction and tangential drive return
toward their reference bands ($e_{\mathrm{unload}}$), release is confirmed and
the program enters \rs{FINAL\_LIFT}; otherwise, it returns to
\rs{PRE\_PRESS\_BASELINE}, acquires a new reference window, and resumes bounded
twisting. The final lift therefore depends on a separate active physical check
rather than on the original release cue alone.

\paragraph{Local execution.}
After compilation, $\Pi$ remains fixed and executes locally. At each control
step, the encoder $E$ converts fresh bilateral tactile observations into
runtime pattern values $\mathbf z_t$. The active supervisor $\sigma_k$
evaluates them together with visual observations, robot state, and execution
memory, and either selects a response in $\mathcal R_k$ or flags a transition.
The program executes the selected response directly as its robot operation, and
$\delta$ updates the active stage once a transition is confirmed. No
foundation-model call occurs in this loop.

\paragraph{Automatic validation and retry.}
Before execution, a schema validator rejects undeclared stages, patterns,
responses, or APIs; out-of-range parameters; unbounded branches; stage
advancement on invalid windows; rotation before press engagement; and
\rs{FINAL\_LIFT} before release verification. A failed check returns its error
to the generation step, and the program is regenerated before the robot trial.

\paragraph{Human intervention.}
A human provides the single RGB-D demonstration and resets the task and sensor
between trials, with randomized task initial states. During execution, the
human neither edits the compiled program nor selects tactile events, responses,
or stage transitions. Stage sequences, selected patterns, response subsets, and
event predicates vary across tasks; the program schema, tactile interface,
bounded-response semantics, validation, and local execution mechanism remain
fixed.

\section{Relation to Prior Work}
\label{app:related_comparison}

Code-as-Monitor~\cite{Zhou_2025_CVPR} compiles visual constraints into local
monitoring code. VLA-Touch~\cite{bi2026vlatouch} refines action segments with a
learned tactile controller. Oller et al.~\cite{Oller_2024} use predefined
desired trajectories and contact modes for tactile closed-loop control. Our
Agent0~+~D comparator, adapted from CaP-X~\cite{pmlr-v306-fu26u}, receives the
same single demonstration and raw tactile stream as CaTS and also executes a
generated program locally. Table~\ref{tab:related_gap_coverage} compares the
task-level challenges addressed by these methods.

\begin{table}[H]
    \centering
    \caption{\textbf{Coverage of contact-rich manipulation challenges.}
    A check denotes explicit support; a cross means the full capability is
    not provided by the cited method or evaluated comparator. CM, A0+D, and
    VLA denote Code-as-Monitor, Agent0~+~D, and VLA-Touch.}
    \label{tab:related_gap_coverage}
    \small\FtabSetup
    \setlength{\tabcolsep}{3pt}
    \begin{tabular}{>{\raggedright\arraybackslash}p{0.40\columnwidth}|c|c|c|c|c}
        \toprule
        \textbf{Challenge} & \textbf{CM} & \textbf{A0+D} &
        \textbf{VLA} & \textbf{Oller} & \textbf{CaTS} \\
        \midrule
        \textbf{Single-demo task strategy} & \xmk & \cmk & \xmk & \xmk & \cmk \\
        \hline
        \textbf{Tactile evidence under occlusion} & \xmk & \cmk & \cmk & \cmk & \cmk \\
        \hline
        \textbf{New-task contact correction and stage-rule synthesis} & \xmk & \xmk & \xmk & \xmk & \cmk \\
        \hline
        \textbf{Task logic reused across sensors} & \xmk & \xmk & \xmk & \xmk & \cmk \\
        \hline
        \textbf{Feedback without per-observation foundation-model calls} & \cmk & \cmk & \cmk & \cmk & \cmk \\
        \bottomrule
    \end{tabular}
\end{table}

CaTS combines a strategy adapted from one demonstration with physical contact
events that select bounded corrections and stage transitions in a local program.
On the child-resistant cap, top-cup extraction, and test-tube task, Agent0~+~D
has 0\% success despite receiving the same demonstration and raw tactile
observations, while CaTS reaches 80\%, 50\%, and 60\%. With the stage plan and
available patterns fixed, demonstration guidance raises next action accuracy
by 11.8 percentage points and closed-loop stage success by 40.0 points. The
cross-sensor tests further show that the same supervision program retains its
event interpretation and overload response through sensor-specific adapters.

\end{document}